\documentclass[11pt]{article}

\usepackage[final]{acl}

\usepackage{times}
\usepackage{subcaption}
\usepackage{latexsym}
\usepackage{xcolor}
\usepackage{amsmath}
\usepackage{amssymb}
\usepackage{hyperref}
\usepackage{booktabs}
\usepackage{graphicx}
\usepackage{multirow}
\usepackage{booktabs}
\usepackage{tabularx}
\usepackage{authblk}
\usepackage{array}
\newcounter{quoteexample}
\newcommand{\quotenum}[1]{%
  \refstepcounter{quoteexample}%
  \label{#1}%
  \hfill(\thequoteexample)%
}

\usepackage{tikz}
\usetikzlibrary{tikzmark}
\usetikzlibrary{patterns.meta}
\usetikzlibrary{shapes.geometric}
\definecolor{CircleBlue}{HTML}{459FE8}
\definecolor{TrianglePink}{HTML}{DE5F88}
\definecolor{SquareGreen}{HTML}{55B96A}
\definecolor{mpblue}{HTML}{1F77B4}
\definecolor{mporange}{HTML}{FF7F0E}

\newcommand{\bluecircle}{%
  \tikz[baseline=-0.6ex]\fill[CircleBlue] (0,0) circle[radius=0.8ex];%
}

\newcommand{\pinktriangle}{%
  \tikz[baseline=-0.6ex]\node[
    regular polygon,
    regular polygon sides=3,
    draw=none,
    fill=TrianglePink,
    minimum size=1.8ex,
    inner sep=0pt
  ] {};
}

\newcommand{\greensquare}{%
  \tikz[baseline=-0.6ex]\fill[SquareGreen] (-0.8ex,-0.8ex) rectangle (0.8ex,0.8ex);%
}
\newcommand{\markword}[2]{%
  \tikz[remember picture,baseline=(#1.base)]%
    \node[
      inner sep=0pt,
      outer sep=0pt,
      anchor=base
    ] (#1) {#2\rule[-0.2ex]{0pt}{4ex}};%
}
\definecolor{object}{HTML}{E39A21}
\definecolor{qbox}{HTML}{459FE8}
\definecolor{pbox}{HTML}{DE5F88}
\definecolor{modelcolor}{HTML}{0076BA}

\DeclareRobustCommand{\gemmabox}{%
  \tikz[baseline=0ex]{
    \draw[draw=black, fill=modelcolor]
      (0,0) rectangle (0.8em,0.8em);
  }%
}

\DeclareRobustCommand{\llamabox}{%
  \tikz[baseline=0ex]{
    \draw[
      draw=black,
      fill=white,
      pattern={Lines[angle=45,distance=2pt,line width=0.4pt]},
      pattern color=modelcolor
    ]
      (0,0) rectangle (0.8em,0.8em);
  }%
}

\usepackage[T1]{fontenc}
\usepackage[utf8]{inputenc}
\usepackage[table]{xcolor}
\newcommand{\hlcell}[1]{\cellcolor{gray!12}{#1}}
\usepackage{microtype}

\usepackage{inconsolata}

\usepackage{graphicx}

\usepackage{authblk}
\usepackage{orcidlink}

\title{A retrieval conditioned rebinding circuit for  dynamic entity tracking in large language models}

\author[1]{Soyoung Oh\,\orcidlink{0000-0002-6466-686X}}
\author[1,2]{Vera Demberg\,\orcidlink{0000-0002-8834-0020}}

\affil[1]{Saarland University}
\affil[2]{Max Planck Institute for Informatics}

\affil[ ]{\texttt{\{soyoung, vera\}@lst.uni-saarland.de}}

\begin{document}
\maketitle
\begin{abstract}
To interpret context correctly and retrieve relevant information, large language models must bind entities to their attributes and update these bindings as state changes. We analyze how LLMs implement this binding process in a dynamic state tracking. Using causal interventions, we identify a \textit{retrieval conditioned rebinding mechanism}, a compact attention head circuit that propagated binding information and when the entity is queried, uses the updated binding to retrieve the corresponding attribute. Across Gemma and Llama models, this circuit supports rebinding behavior, but the representational signature of the mechanism differs across model families. In Gemma models, the binding signature is clearly expressed in the query/key subspaces of the relevant attention heads, whereas in Llama models, the binding information is carried primarily in key vectors. Overall, our results reveal an interpretable mechanism for context dependent state tracking in LLMs\footnote{The code is publicly available at \url{https://github.com/sori424/dynamicET}.}.
\end{abstract}

\section{Introduction}

Accurately maintaining entity specific information over extended contexts is a fundamental requirement for a long context language understanding. Prior work has shown that large language models can succeed at entity tracking, yet their behavior remains inconsistent across settings~\cite{kim2023entity, li2022emergent, kim2024code}. This raises a question about the nature of the underlying computation. Although recent studies have probed the mechanisms behind entity tracking, they have not fully examined settings that require explicit state transitions~\cite{prakash2024fine} or naturalistic language contexts~\cite{li2025language}. In this work, we study the mechanism for an entity tracking in a natural language setting with dynamic state updates.

We frame this problem through the lens of binding~\cite{feng2024language, feng2025monitoring}, a representational capacity that supports compositional reasoning~\cite{fodor1988connectionism}. The binding refers to a process of associating the features of an object with the object itself, while keeping those features distinct from those of other objects~\cite{treisman1996binding}. Our interest is not only whether language models can form such associations, but also whether they can update them when the underlying state changes.

% do models genuinely maintain structured representations of entities and their attributes, or do they rely on superficial, task-specific heuristics that are brittle to ablations? 

% This distinction is important because successful performance may arise either from maintaining and updating latent entity--attribute bindings or from applying shortcut strategies that only approximate the correct answer at query time.

%%%%%%%%%% TODO: ADD THE BINDING-ID PART
Consider a simple example involving three boxes and three objects, followed by a state update which is adapted from~\citet{kim2023entity}:

\begin{quote}
Context: \markword{boxx1}{\textcolor{qbox}{Box R}} contains the \markword{milk}{\textcolor{qbox}{rabbit}}.
\markword{boxp1}{\textcolor{pbox}{Box S}} contains the \markword{cup1}{\textcolor{pbox}{sock}}.
\markword{boxu1}{\textcolor{SquareGreen}{Box T}} contains the \markword{egg1}{\textcolor{SquareGreen}{toy}}.
Swap the items of \markword{boxp2}{\textcolor{pbox}{Box S}} and \markword{boxx2}{\textcolor{qbox}{Box R}}.

Question: Which item does \textbf{Box R} contain?

Answer: \markword{cup}{\textcolor{object}{sock}}\quotenum{quote:example}

\begin{tikzpicture}[remember picture,overlay]
  % blue circles
  \foreach \n in {boxx1,milk,boxx2}{
    \fill[CircleBlue] ([yshift=-1ex]\n.south) circle[radius=0.7ex];
  }

  % pink triangles
  \foreach \n in {boxp1,cup,cup1,boxp2}{
    \node[
      regular polygon,
      regular polygon sides=3,
      draw=none,
      fill=TrianglePink,
      minimum size=2ex,
      inner sep=0pt
    ] at ([yshift=-1ex]\n.south) {};
  }

  % green squares
  \foreach \n in {boxu1,egg1}{
    \node[
      draw=none,
      fill=SquareGreen,
      minimum size=1.4ex,
      inner sep=0pt
    ] at ([yshift=-1ex]\n.south) {};
  }

%%% Blue to Pink
  %   \node[
  %   circle,
  %   draw=none,
  %   fill=CircleBlue,
  %   minimum size=1.6ex,
  %   inner sep=0pt
  % ] (cir) at ([yshift=-1.0ex,xshift=-2ex]qboxx.south) {};

  % same pink triangle style as before
  % \node[
  %   regular polygon,
  %   regular polygon sides=3,
  %   draw=none,
  %   fill=TrianglePink,
  %   minimum size=2.0ex,
  %   inner sep=0pt
  % ] (tri) at ([yshift=-1.0ex,xshift=2ex]qboxx.south) {};

  % \draw[->,thick] ([xshift=0.3ex]cir.east) -- ([xshift=-0.3ex]tri.west);
\end{tikzpicture}
\end{quote}

To answer this question correctly, a model must first represent the initial assignments \texttt{contains(R,rabbit)}, \texttt{contains(S,sock)}, and \texttt{contains(T,toy)}. \citet{feng2024language} showed that the model solves this \emph{binding problem} by relying on binding IDs (\bluecircle, \pinktriangle, \greensquare), represented as vectors that are added to lexical information in the activation space. A natural hypothesis for how the model handles the state update is a \textbf{Global State Update}. Under this account, the model fully re-encodes the post update state immediately after the swap operation, analogous to mental simulation in cognitive accounts of language comprehension and reasoning~\cite{johnson1983mental}. On this view, the model maintains a situation model of the described world and incrementally updates it at the swap into a complete post swap configuration. For example, right after the \texttt{Swap} operation, the latent context would encode updated relations such as \texttt{contains(R,sock)} and \texttt{contains(S,rabbit)}. Answering then requires only reading off the object currently bound to the referenced box in this updated latent state.

An alternative hypothesis is the \textbf{Retrieval Conditioned Rebinding}. Under this account, the model does not re-encode the full situation immediately after each state transition. Instead, the model may preserve compact swap related information and apply it only during answer retrieval. Under the binding framework, therefore, a \texttt{Swap} operation need not permute all box--object bindings. When the final question asks about \texttt{Box R}, the model would use the locally updated binding ID for \texttt{Box R}, e.g., from $\bluecircle$ to $\pinktriangle$, and then use $\pinktriangle$ to retrieve the associated object, \textit{sock}. The answer is therefore obtained through a local rebinding followed by binding ID based readout, rather than by reconstructing a globally updated post swap state of the entire context.

% These two accounts make different mechanistic predictions. If models rely on full-state re-encoding, then representations of the affected boxes should change in content-like ways after the update, and interventions on the post-swap representations should directly alter the answer. If models instead rely on pointer-like efficient mapping, then the relevant computation should be localized to the binding IDs or mapping variables that mediate the relation between entities and fillers. In that case, the lexical representation of an object may remain relatively stable, while the answer depends on a dynamically updated assignment structure. 

We investigate this question using the dynamic state tracking task~\cite{kim2023entity} as in Example~\ref{quote:example}, together with causal interventions on LLM activations. The analysis proceeds in four steps and is run on four instruction-tuned models \{Gemma2-9B, Gemma3-12B, Llama3.2-3B, and Llama3.1-8B\} to test generalizability of the circuit: 

\begin{enumerate}
    \item We use causal mediation analysis~\cite{prakash2025belief, lieberum2023does, gur2025mixing} to localize the token level routes through which role specific information influences the final prediction. We observe that the target object is retrieved directly from its original contextual occurrence, whereas information about the boxes are mediated through multiple intermediate token positions. This dissociation supports a \textit{retrieval conditioned rebinding}, where swap-relevant box information is transferred through intermediate positions and applied during readout time, rather than stored as a globally updated box--object state. 
    
    \item Next, we refine these coarse routes using path patching~\cite{wang2023interpretability, prakash2024fine}. This reveals a compact attention head circuit, which consists of five functional groups, each associated with token positions identified in the previous step. The circuit, comprising 3--10\% of the full model components, recovers near full model performance on the task.
    
    \item To discern the role of each attention head, we conduct role specific interchange intervention~\cite{vig2020investigating, prakash2025belief, geiger2020neural}. We find that one group functions as a pointer, which is sensitive to address selection, suggesting that these heads mediate binding identifier routing.
    
    \item To test whether this binding identifier is causally involved in predicting the final answer, we adapt the binding intervention~\cite{feng2024language} to the subspaces of the relevant attention head group. We observe that modifying the binding identifier redirects the model's attention to the corresponding object and shifts the final logits accordingly, but with different signatures across model families.

% We observe that modifying the binding identifier redirects the model's attention to the corresponding object and shifts the final logits accordingly, but with different signatures across model families. Gemma models exhibit a clearer attention shift to the newly indicated object, whereas Llama models show a weaker or more distributed redirection pattern.
    
\end{enumerate}

\section{Task formulation and rebinding hypothesis}In the dynamic entity tracking task, LLMs represent an abstract binding ID shared by each box and its associated object, analogous to the binding mechanism proposed by \citet{feng2024language, feng2025monitoring}. Under this account, successful reasoning depends on preserving and updating box--object relations at the level of binding IDs, rather than at the level of surface token identity.

Let $\mathcal{B}$ denote a set of boxes and $\mathcal{O}$ a set of objects. A task instance of size $n$ consists of distinct boxes $B_0,\dots,B_{n-1}\in\mathcal{B}$ and distinct objects $O_0,\dots,O_{n-1}\in\mathcal{O}$, arranged into an initial context $c_0 = \mathrm{ctxt}(B_0 \mapsto O_0,\; \dots,\; B_{n-1} \mapsto O_{n-1})$. So, in Example~\ref{quote:example}, this corresponds to $c_0 = \mathrm{ctxt}(\mathit{R} \mapsto \mathit{rabbit},\; \mathit{S} \mapsto \mathit{sock},\; \mathit{T} \mapsto \mathit{toy})$.

\paragraph{Binding mechanism.}
We assume that each initial pair $(B_k,O_k)$ is associated with a shared binding ID $k$. The box representation encodes the box identity together with this binding ID, and the object representation encodes the object identity together with the same ID. We denote these representations as $Z_{B_k} = \Gamma_B(B_k,k),\quad Z_{O_k} = \Gamma_O(O_k,k)$.

Under this view, answering a question about $B_k$ requires retrieving the object whose binding ID matches the one assigned to $B_k$.

\paragraph{Rebinding mechanism.} The swap operation $\operatorname{swap}(i,j)$ exchanges the objects associated with boxes $B_i$ and $B_j$. A natural interpretation for this transition is a \textit{global state update}, where the model constructs a complete post swap context right after the change. For example, applying $\mathrm{swap}(0,1)$ in Example~\ref{quote:example} yields updated context $c_1 = \mathrm{ctxt}(\mathit{R} \mapsto \mathit{sock},\; \mathit{S} \mapsto \mathit{rabbit},\; \mathit{T} \mapsto \mathit{toy})$.

Under this account, the model answers by reading off the object bound to the referenced box in the updated latent state. On the other hand, the model may not construct the entire swapped world state in advance. Instead, it may preserves the original bindings and reinterpret them only when retrieval demands it. This motivates a more targeted alternative, the \textit{retrieval conditioned rebinding} hypothesis. Under this account, the model preserves the original box--object representations from $c_0$ and applies the swap as a remapping over binding IDs during answer retrieval. We formalize this remapping as
\[
\rho_{i,j}(k)=
\begin{cases}
j & \text{if } k=i,\\
i & \text{if } k=j,\\
k & \text{otherwise}.
\end{cases}
\]
Given a question about box $B_k$, the model applies the readout permutation $\rho_{i,j}$ to the referred binding ID, yielding $k'=\rho_{i,j}(k)$, and then retrieves the object representation whose stored binding ID matches $k'$, $\Gamma_O(O_{k'},k')$.

In Example~\ref{quote:example}, the referenced box is $\mathit{R}$, whose original binding ID is $0$. After $\mathrm{swap}(0,1)$, retrieval conditioned rebinding maps this ID to $1$, allowing the model to retrieve $\Gamma_O(\mathit{sock},1)$ from the original context and return $\mathit{sock}$. To test these hypotheses, we start with causal mediation analysis to trace how information flows to the final answer position.
\section{Tracing information flow of crucial input tokens}

We focus on three functionally distinct token roles: the reference box (\textcolor{qbox}{Box R}), the swap target box (\textcolor{pbox}{Box S}), and the target object (\textcolor{object}{sock}). We conduct three separate intervention experiments: \textcolor{qbox}{Exp 1} tests how information associated with the reference box flows by combining contextual content with the swap operation; \textcolor{pbox}{Exp 2} tests how information from the swap target box is transferred to the question; \textcolor{object}{Exp 3} reveals how the object value is first instantiated and then routed through to determine the answer.  

%%%% [TODO]: keep the base same for the all experiments in the Figure 1 and only change the color for the text that being patched (change the box into R,S,T -- rabbit, ...).

\begin{figure*}[!h]
    \centering
    \includegraphics[width=0.8\textwidth]{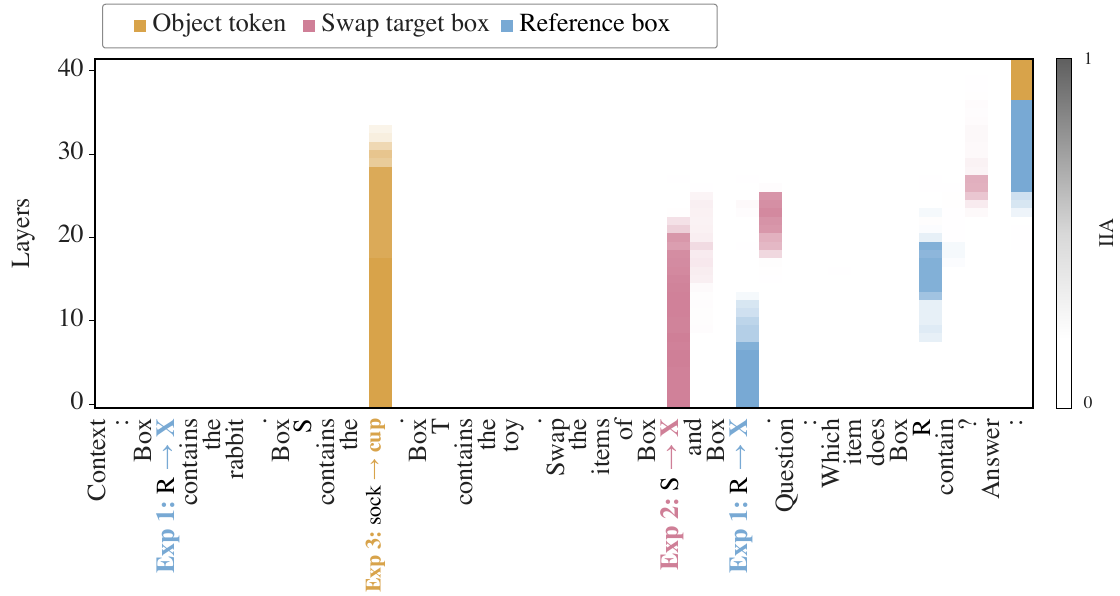}
    \caption{Information flow of crucial tokens using interchange interventions in Gemma-9B. \textbf{y-axis:} model layers; \textbf{x-axis:} shared original prompt template. Each colored token is varied in a separate experiment: \textcolor{qbox}{Exp1. Reference box}, \textcolor{pbox}{Exp2. Swap target box}, \textcolor{object}{Exp 3. Object}, all other black tokens are unchanged; \textbf{cell:} reports normalized IIA, darker cells indicating stronger causal mediation.}
    \label{fig:causal_mediation}
\end{figure*}

\paragraph{Causal mediation analysis.}
%% TODO: Explain more about the sub experiments.
To trace how answer-relevant information carried by role-bearing tokens flows into the final prediction, we conduct a coarse grained information flow analysis by using interchange interventions~\cite{prakash2025belief, geiger2021causal}. For each role, we construct an original input $x_{\mathrm{orig}}$ and a counterfactual input $x_{\mathrm{counter}}$ that differ only in the role-bearing tokens relevant to that experiment. We then patch residual stream activations from $x_{\mathrm{orig}}$ into the corresponding token layer position of $x_{\mathrm{counter}}$. For example, patching from an original context such as \texttt{Box S contains the sock} into a counterfactual context such as \texttt{Box S contains the cup} tests whether the patched activation is sufficient to shift the model from the counterfactual answer behavior, e.g., \texttt{cup}, toward the original expected answer, e.g., \texttt{sock}. We measure the interchange intervention accuracy (IIA), which indicates the extent to which patching that position causes the model to recover the original answer behavior~\cite{geiger2022inducing}. A higher IIA means that position is a candidate for carrying causal signal for the answer prediction.

% We replace the counterfactual input's (e.g., $x_{counter}$: Box X contains the cup.) residual stream vectors with those computed from the original input (e.g., $x_{orig}$: Box R contains the rabbit.) so that by patching, the expected output changes from \textit{unknown/cup} to \textit{sock}. We measure the interchange intervention accuracy (IIA), which is the indicator for how successfully patching that position causes the model to realize the expected counterfactual answer behavior~\cite{geiger2022inducing}. A higher IIA means that position is a candidate for carrying causal signal for the answer prediction.

% \begin{figure*}[htbp]
%     \centering
%     \includegraphics[width=0.7\textwidth]{./figures/iia.pdf}
%     \caption{Information flow of crucial tokens using causal mediation analysis.}
%     \label{fig:causal_mediation}
% \end{figure*}

\paragraph{Results.}Figure~\ref{fig:causal_mediation} shows the aggregated result of the causal mediation analysis (per-experiment results in Appendix Figure~\ref{fig:cm}) over $N=300$ examples. The effects indicate where information associated with each role contributes to the final answer prediction. Information associated with the reference box (\textcolor{qbox}{Box R}) and swap target box (\textcolor{pbox}{Box S}) does not appear to be routed from their initial contextual mentions. Instead, their causal effects emerge mainly after the \texttt{Swap} operation and propagate toward the final prediction position. This pattern suggests that the model does not answer by directly retrieving the swap target box's original contextual content. Consistent with this, patching the readout position (`:') has a stronger causal effect than patching the swap operation position (`.' token in the swap sentence), suggesting that the final binding is resolved primarily during retrieval (Figure~\ref{fig:cm_additional}). 

In contrast, information about the answer object (\textcolor{object}{sock}) is routed more directly. The object token's residual stream contributes to the final prediction in later layers, rather than being mediated through the box label tokens. This aligns with prior findings~\cite{prakash2025belief, prakash2024fine, lieberum2023does, gur2025mixing}. 

Together, this pattern contrasts with what we would expect under a \textit{global state rebinding}. If the model reconstructed a full post swap state, the answer object should be mediated through the updated box representation after the \texttt{Swap}. The fact that this is not the case suggests that the model does not rely primarily on a globally updated box--object representation. Rather, the observed pattern supports the \textit{retrieval conditioned rebinding}, as the model appears to compute the relevant post swap box relation separately from object retrieval. That is, the \texttt{Swap} operation mediates a rebinding from the referenced box to its swapped target which then guides retrieval of the answer object from its original contextual position in later layers. This finding aligns with \citet{tang2026language}, who argue that LMs do not incrementally track state updates across tokens but instead aggregate relevant information in parallel once the query becomes available. We further validate this finding in more complex settings by increasing the number of boxes and an additional distractor swap where the results remain consistent (Appendix~\ref{sec:generalization-threebox}).

\section{Localizing the rebinding mechanism}
The causal mediation analysis provides token level evidence for a \textit{retrieval conditioned rebinding} mechanism. To verify this mechanism at the attention head level, we next apply path patching~\cite{wang2023interpretability, prakash2024fine}.

\subsection{Identifying circuit components with path patching}
\label{sec:path-patching}
Concretely, for each example ($N=100$) we construct a clean input $x_{\mathrm{orig}}$ and a corrupted input $x_{\mathrm{noise}}$, obtained by randomizing the box--object assignments while preserving the prompt structure. Let $O_k$ be the correct answer for $x_{\mathrm{orig}}$. We compute the clean probability \(p_{\mathrm{orig}}=p_M(O_k\mid x_{\mathrm{orig}})\) and the patched probability \(p_{\mathrm{patch}}=p_{M_{\mathrm{patch}}}(O_k\mid x_{\mathrm{orig}})\), where the patched activation comes from \(x_{\mathrm{noise}}\). We add the paths with the most negative \((p_{\mathrm{patch}}-p_{\mathrm{orig}})/p_{\mathrm{orig}}\) at each selection iteration. 

\begin{figure*}[!h]
    \centering
    \includegraphics[width=0.9\textwidth]{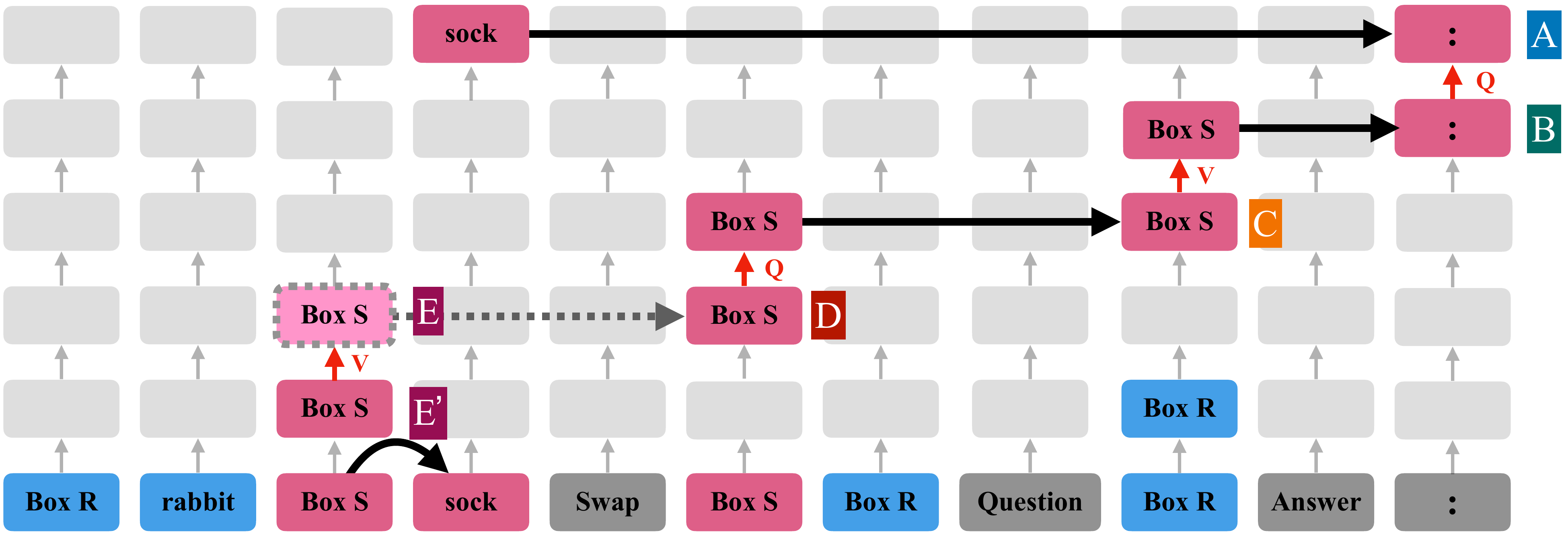}
    \caption{Path patching circuit for the retrieval conditioned rebinding mechanism. Rows denote functional head groups A--E at specific tokens; Columns denote layers. Horizontal arrows indicate attention mediated value routing. Vertical dashed arrows indicate residual stream composition between head groups, with V and Q denoting value and query mediated composition. Information flows via the chain $(\text{E} \xrightarrow{V})~\text{D} \xrightarrow{Q} \text{C} \xrightarrow{V} \text{B} \xrightarrow{Q} \text{A} \rightarrow \text{logit}$. E$'$ marks the context side box--object binding formation, while E marks a later binding anchor representation that writes the binding information into the readout circuit. The dotted Group E path is model dependent: it is active in some models, but not in Gemma-9B after pruning -- binding anchor information may be routed by other heads or already available in the destination position's residual stream at readout.}
    \label{fig:path}
\end{figure*}

% Here, we aim to identify five groups of attention heads, each responsible for a distinct sub-computation: (A) Answer retriever (B) Dereferencer (C) Position updater (D) Swap position transmitter (E) Binding anchor (Figure~\ref{fig:path}). 

% The circuit's computation proceeds from Group E up to Group A. Group E, the structural reader at the context-target box position, reads the box-identity information from the token embedding and writes it into the residual stream. Group D, the query modulator at the target-box position, shapes the attention query that guides the queried-box token toward the appropriate earlier position. Group C, the position detector at the question-queried-box position, uses this query-side modulation to attend to that position and recover the updated box identity. Group B, the position transmitter, attends from the final token position back to the question-queried-box position and reads Group C's output via V-composition, relaying the resolved box-position information into the residual stream at the final token. Finally, Group A consists of value fetcher heads — identified as the heads with the lowest patching score on the correct answer logit at the final token position — which use the representation deposited by Group B through Q-composition to form an attention pattern over the correct object token (e.g., `cup'), thereby producing the correct answer.

We use two forms of path patching to distinguish two forms of sender--receiver interaction: \textit{V-patching} and \textit{Q-patching} (marked as \textcolor{red}{V, Q} in Figure~\ref{fig:path}). In V-patching, we replace the value side input to a receiver head to test whether it reads information written by an upstream sender at the attended source position. In Q-patching, we replace the query side input at the receiver head's target position to test whether an upstream sender shapes the receiver's query, thereby changing its attention pattern.

\paragraph{Rebinding circuit.}
% We use the five group circuit as an interpretive scaffold (Figure~\ref{fig:path}), adapted from~\citet{prakash2024fine}, each responsible for a distinct computation in the rebinding circuit: 
% We use the five functional roles proposed by \citet{prakash2024fine} as an interpretive scaffold for analyzing rebinding behavior, rather than assuming that the same circuit is reused in our setting. These roles provide possible subcomputations in the circuit: (A) answer retrieval, (B) dereferencing, (C) position updating, (D) swap-position transmission, and (E) binding anchoring. In our analysis, however, the heads are selected independently by causal pruning on our task and model, and are subsequently compared against this functional decomposition. Thus, the grouping serves as a descriptive framework for interpreting the discovered circuit, not as a pre-specified circuit imported from prior work.

We use the five functional roles employing \citet{prakash2024fine} as an interpretive scaffold, not as a pre-specified circuit. These roles describe candidate rebinding subcomputations (Figure~\ref{fig:path}): (A) Answer retriever, (B) Dereferencer, (C) Position updater, (D) Swap position transmitter, and (E) Binding anchor. The heads in our circuit are selected independently through causal pruning on our task and model. The rebinding circuit computation proceeds from Group E to Group A. Group E acts as a binding anchor at the context target box position, writing box identity information into the residual stream. Group D, the swap position transmitter, reads this information through V-composition and propagates it along the swap path. Group C, the position updater, receives information via Q-composition, which shapes Group C's attention toward the appropriate earlier box position to the updated binding. Group B, the dereferencer, then reads the resolved binding information through V-composition. Finally, Group A consists of answer retriever heads at the final position that representation supplied by Group B shapes Group A's attention over target object token, increasing the logit of the correct answer (i.e., \texttt{sock}). We used a five stage greedy path patching process. In each stage, all attention heads were scored as candidate senders over 100 prompts; the top-k heads with the most negative mean relative target probability change were retained. The fixed top-k values were 50,50,25,10, and 10 for Groups A-E respectively following the logic of the prior work~\cite{prakash2024fine}, and the procedure stopped after Group E with no early stopping.

\paragraph{Circuit evaluation.}We evaluate the discovered rebinding circuit using candidate accuracy where a prediction is counted as correct if the gold object $O_k$ receives the highest logit among the candidates. 

Let $F$ denote candidate accuracy averaged over $N=300$ held-out examples, disjoint from the examples used to identify the circuit. We define $F(M)$ for the corresponding full model candidate accuracy and $F(Cir)$ as the circuit accuracy. We also set a random baseline by sampling 10 random subsets of attention heads with the same size as the candidate circuit and applying the same mean ablation procedure. We prune the candidate circuit using a contribution based criterion~\cite{wang2023interpretability}. For each head $v \in Cir$, and a subset of other circuit heads $K \subseteq Cir \setminus \{v\}$, we define the contribution of $v$ as $(F(Cir \setminus K)-F(Cir \setminus (K \cup \{v\})))/F(Cir \setminus (K \cup \{v\}))$. Heads with contribution score less than $1\%$ are pruned. 

\paragraph{Results.}We find that only 38 out of 672 attention heads (5.7\%), are sufficient to recover most of the model's rebinding behavior. The circuit reaches 0.89 accuracy compared with 1.00 for the full model and 0.34 for a random head baseline, indicating that the relevant computation is localized in a compact attention head circuit. The selected heads and additional evaluation metrics are reported in Appendix~\ref{sec:circuit-heads}. Notably, the pruned circuit retains no heads from Group E. Since pruning removes attention head outputs but preserves the residual stream, binding relevant information may already be available in the relevant token positions due to earlier context side binding formation, marked as E$'$ in Figure~\ref{fig:path}, and then carried forward through the residual stream.
% Alternatively, the model may not construct an explicit fully swapped object representation at all, but instead resolve the relevant binding only at retrieval. Thus, the absence of Group E heads suggests that this stage is either redundant under the pruning intervention, implemented through components outside our hand-defined group, or unnecessary under a retrieval-conditioned rebinding strategy.

% After pruning, the model retains no heads in Group E, suggesting that the functional decomposition may be implemented redundantly or through model specific variants.

% As shown in Table~\ref{tab:entity_tracking_acc}, the pruned circuits achieve accuracy close to the full models and substantially above random same-size subsets. Gemma-9B and Gemma-12B reach circuit accuracies of 0.89 and 0.91, compared with random baselines of 0.34 and 0.38. Llama-8B reaches 0.84 versus 0.38 random accuracy, and Llama-3B matches the full model accuracy of 0.71. Thus, a small set of selected heads (3\%--10\% across models) preserves much of the task performance. The selected heads and additional evaluation metrics are reported in Appendix~\ref{sec:circuit-heads}.

% model_metrics.candidate_accuracy
% minimal_circuit_metrics

\subsection{Identifying address retrieval routing head group}
\label{sec:retrieval-routing-group}
% https://arxiv.org/pdf/2505.14685
The previous path patching results suggest that at `:' position, the model first determines the relevant source position through attention routing, and then retrieves the object information stored at that position. This resembles the lookback mechanism of \citet{prakash2025belief}, where retrieval involves matching a later pointer token to an earlier address token, so that the model attends to the earlier token and reads out the associated payload. We therefore apply interchange interventions~\cite{vig2020investigating, geiger2020neural} to the attention edges of each head group, testing whether its effect is carried by edges that select the retrieval address.

\begin{figure}[!h]
    \centering
    \begin{subfigure}[t]{\columnwidth}
        \centering
        \includegraphics[width=\linewidth]{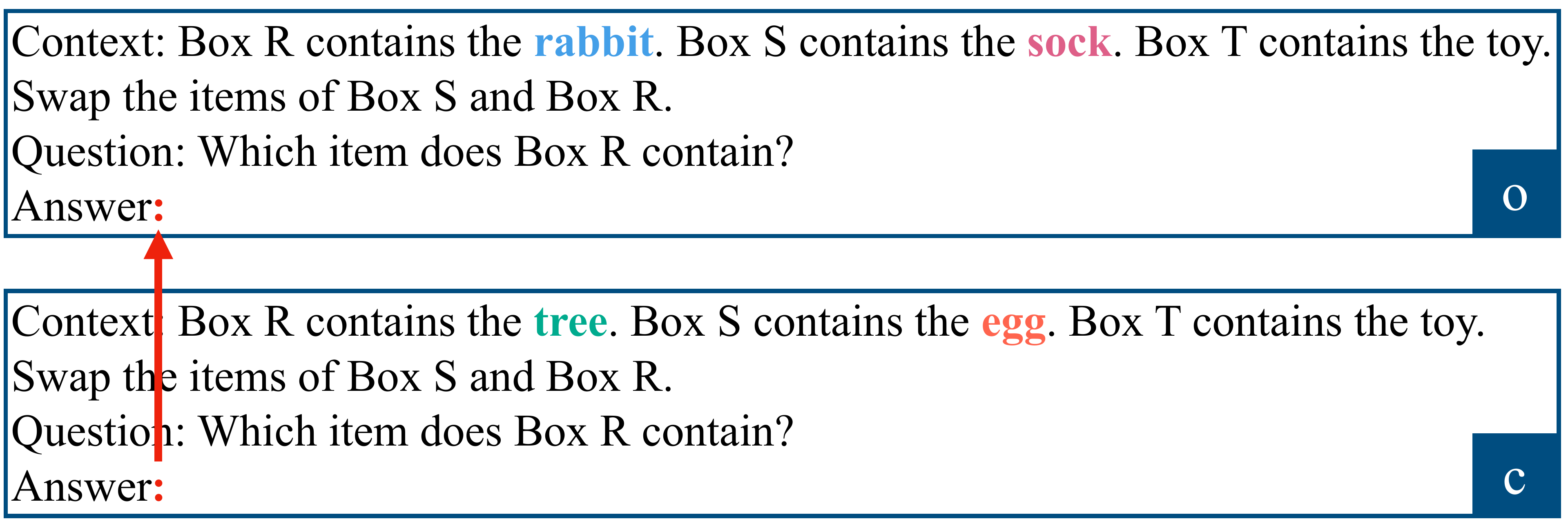}
        \caption{\textbf{Content control intervention:} If the heads mediate address routing rather than object content transfer, attention pattern patching should not increase the counterfactual object logit, so we expect $\Delta \mathrm{logit}(\texttt{egg}) \approx 0$.} 
        \label{fig:payload}
    \end{subfigure}
      \begin{subfigure}[t]{\columnwidth}
        \centering
        \includegraphics[width=\linewidth]{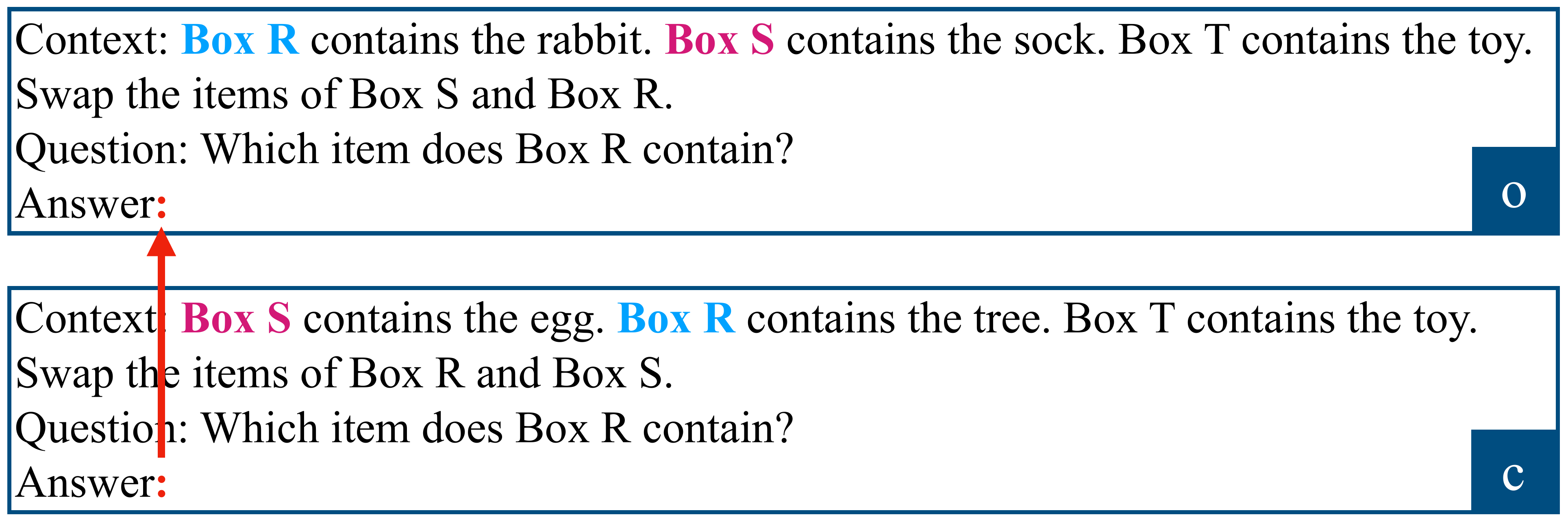}
        \caption{\textbf{Pointer intervention:} If the heads mediate retrieval addressing, we expect $\Delta \mathrm{logit}(\texttt{rabbit}) > 0$, where \texttt{rabbit} is the token at the counterfactual position $0$ in the original prompt.}
        \label{fig:pointer}
    \end{subfigure}

    \caption{Attention pattern interchange paired prompts: o (original), c (counterfactual).}
    \label{fig:attn_pattern_roles}
\end{figure}

\paragraph{Patch construction and target logits.}
We construct paired original and counterfactual inputs ($N=300$) and patch the attention patterns of the selected heads from the counterfactual run into the corresponding heads at the `:' position of the original run. We then measure the change in the logit of object token $O_k$, $\Delta \mathrm{logit}=\mathrm{logit}(O_k)_{\mathrm{patched}}-\mathrm{logit}(O_k)_{\mathrm{original}}$. We use two variants. In the \textit{content control intervention} (Figure~\ref{fig:payload}), the object tokens are changed while the box labels and swap structure are kept fixed. Since the retrieval address is unchanged, e.g., (\textcolor{qbox}{Box R, 0}) and (\textcolor{pbox}{Box S, 1}), patching only the attention pattern should continue to route retrieval to the same source position in the original prompt, rather than transferring the counterfactual object token. Thus, $\Delta \mathrm{logit}(\texttt{egg})$ serves as a control for whether attention pattern patching alone transfers counterfactual object content independently of address selection.

In the \textit{pointer intervention} (Figure~\ref{fig:pointer}), we change the binding structure, (\textcolor{pbox}{Box S, 0}), (\textcolor{qbox}{Box R, 1}), so that the counterfactual prompt requires retrieval from a different source position. If a head group controls address selection, then patching its counterfactual attention pattern into the original run should shift the model toward the object located at the counterfactual source position in the original prompt (i.e., \textcolor{qbox}{\texttt{rabbit}}, the object stored at binding ID \(0\) in the original prompt).

\begin{figure}[!h]
    \centering
    \includegraphics[width=0.4\textwidth]{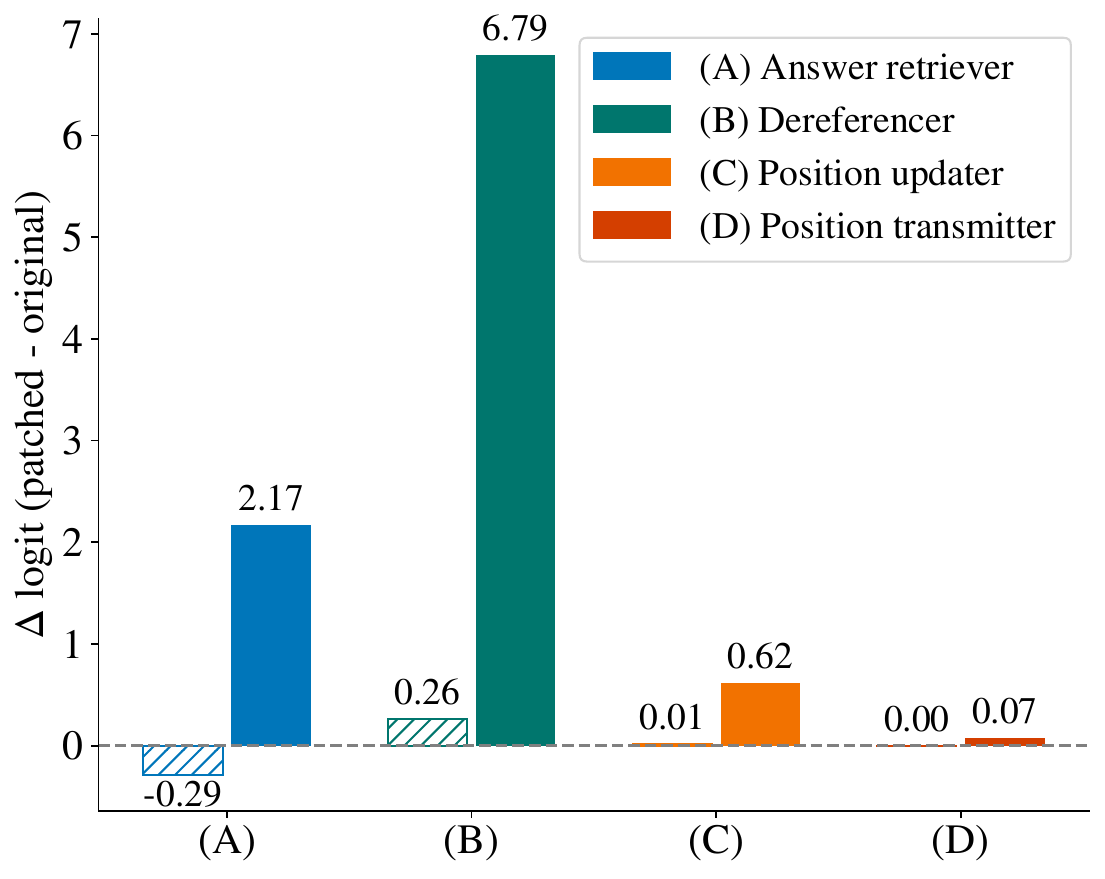}
 \caption{Role specific head patching effects for Gemma-9B. 
\gemmabox~$\Delta \mathrm{logit}(\texttt{rabbit})$ is represented by fully colored bars, while \llamabox~$\Delta \mathrm{logit}(\texttt{egg})$ is represented by diagonally hatched bars.}
    \label{fig:role_attn_head}
\end{figure}

\paragraph{Results.}As shown in Figure~\ref{fig:role_attn_head}, in the pointer intervention, patching the counterfactual attention pattern for Group B substantially increases $\Delta \mathrm{logit(\texttt{rabbit})}=6.79$. In contrast, the same group produces only a small change in the content control target $\Delta \mathrm{logit(\texttt{egg})}=0.26$. This dissociation indicates that Group B primarily mediates retrieval routing through address selection, rather than transferring counterfactual object content. Group A, the answer retriever heads, also increases the pointer target logit ($\Delta \mathrm{logit}(\texttt{rabbit})=2.17$), while decreasing the content control target logit ($\Delta \mathrm{logit}(\texttt{egg})=-0.29$). This suggests that Group A may also participate in routing to the retrieved source position. Overall, the results identify Group B as the clearest address mediated retrieval routing group, with weaker contributions from Groups A and C\footnote{We exclude Group E because no heads satisfy the pruning criterion.}.

\section{Probing binding ID matching in Q/K subspaces of dereferencer heads}
\label{sec:bid-intervention}
The attention pattern interchange results show that Group B plays a causal role in redirecting retrieval across binding positions. Therefore, we ask whether Group B's pointer role arises from a binding ID comparison in its Q/K subspaces. Since a head's attention scores are computed by query--key dot products, Group B could implement retrieval routing by binding ID $k$ at `:' position, while keys at earlier context tokens encode corresponding binding ID $k$. Matching binding IDs would then cause the head to attend to the corresponding answer object token. We test this with binding ID interventions in Q/K space of Group B heads.

Following the additive representation hypothesis~\cite{elhage2021mathematical} and the binding ID intervention framework~\cite{feng2024language}, we treat
binding IDs as approximately linearly represented directions in the activation space. We estimate a binding ID direction in the query and key spaces of the selected Group B heads. For a head $h$ in layer $\ell$, let
\[
A_{\ell,h}(t,s)
=
\mathrm{softmax}
\left(
\frac{
q_{\ell,h}(t)^\top k_{\ell,h}(s)
}{
\sqrt{d_h}
}
\right)
\]
denote the attention pattern from a target position $t$ to a source position $s$. In our task, the target position $t_{\mathrm{:}}$ is the `:' token, and the source positions are the context tokens corresponding to box and object tokens.

\paragraph{Binding ID directions in Q/K subspace.}We estimate binding ID directions by averaging Group B query and key vectors over $N=100$ prompts with randomized lexical items, so that lexical variation is averaged out. Let $r(x)$ denote the retrieval binding ID in prompt $x$. For each binding ID $k$, we compute the mean query vector at the `:' position as $\mu^Q_{\ell,h}(k)=
\mathbb{E}_{x:\, r(x)=k}
\left[
q^x_{\ell,h}(t_{\mathrm{:}})
\right]$. For the key side, let $S_k(x)$ denote the set of context box/object token positions carrying binding ID $k$ in prompt $x$, then
$\mu^K_{\ell,h}(k)
=
\mathbb{E}_{x}
\left[
\frac{1}{|S_k(x)|}
\sum_{s\in S_k(x)}
k^x_{\ell,h}(s)
\right]$.

\paragraph{Q/K intervention.}We define shift directions as differences between the mean Q/K representations for two binding IDs:
\[
\begin{aligned}
\Delta^Q_{\ell,h}(i \rightarrow j)
=
\mu^Q_{\ell,h}(j) - \mu^Q_{\ell,h}(i),
\\
\Delta^K_{\ell,h}(i \rightarrow j)
=
\mu^K_{\ell,h}(j) - \mu^K_{\ell,h}(i).
\end{aligned}
\]
Assuming that binding ID is approximately linearly represented, \(\mu(j)-\mu(i)\) estimates the representational shift from ID \(i\) to ID \(j\). We use these directions for causal interventions in the query and key subspaces. For each original prompt requiring retrieval from binding ID $i$, we pair a counterfactual binding ID $j$ and apply three interventions.

In the \textit{query-side intervention}, we shift the query vector at the `:' position from ID ($i$) to ($j$) by
\[
q_{\ell,h}(t_{\mathrm{:}})
\leftarrow
q_{\ell,h}(t_{\mathrm{:}})
+
\Delta^Q_{\ell,h}(i \rightarrow j),
\]
This tests whether changing the binding ID requested at the answer position redirects Group B attention from source positions carrying ID $i$ to $j$.

In the \textit{key-side intervention}, we keep the query fixed but change the binding IDs represented by the context tokens at box only, object only, and both positions. Let $S_k^r$ denote the selected context positions associated binding ID $k$, where $r \in \{\mathrm{box}, \mathrm{object}, \mathrm{both}\}$. We then apply
\[
\begin{aligned}
k_{\ell,h}(s)
\leftarrow
k_{\ell,h}(s)
+
\Delta^K_{\ell,h}(j \rightarrow i),
~s \in S_j^r,
\\
k_{\ell,h}(s)
\leftarrow
k_{\ell,h}(s)
+
\Delta^K_{\ell,h}(i \rightarrow j),
~s \in S_i^r.
\end{aligned}
\]
If Group B attends by Q/K binding ID matching, the unchanged ID $i$ query should now match the shifted keys at $S_j^r$, redirecting attention to the counterfactual context. The three variants test whether this match is carried by keys of box, object, or both.

In the \textit{Q+K intervention}, we apply both interventions together. The query is shifted from ID $i$ toward $j$, while the keys at the original source positions $S_i$ are also shifted from ID $i$ toward $j$. Thus, if Group B computes the pointer through Q/K compatibility, the shifted query should again match the original source positions $S_i$, now via their shifted keys. Thus, the combined intervention should cancel the redirection observed in the Q-only and K-only interventions.

% Concretely, consider a prompt with
% \[
% \text{$k=$ 0: Box X} \rightarrow \text{milk},
% ~
% \text{$k=$ 1: Box P} \rightarrow \text{cup},
% \]
% followed by \texttt{Swap X and P} and the query \texttt{Box X}. The correct
% post-swap source is slot 1, so the correct answer is \texttt{cup}. If we choose
% slot 0 as the target source, the Q-side intervention shifts the answer-position
% query from source ID 1 to source ID 0, and the K-side intervention swaps the
% source-side IDs of slots 0 and 1. Either intervention should redirect Group B's
% attention from slot 1 to slot 0, moving the final logits toward
% \texttt{milk}. Applying both interventions together should restore compatibility
% between the shifted query and the original source slot, moving attention and
% logits back toward slot 1 and \texttt{cup}.

\paragraph{Metrics.}We report three metrics averaged over $N=300$ examples: (i) $\Delta R$ measures how much the intervention shifts attention from context tokens associated with the original binding ID $i$ toward counterfactual binding ID $j$; positive values indicate a shift toward the counterfactual binding. (ii) $\Delta$logit measures the corresponding change in the final output logit margin favoring the counterfactual answer over the original answer. (iii) Switch fraction measures how often the intervention changes the dominant attended binding ID from $i$ to $j$. Formal definitions of these metrics are provided in Appendix~\ref{sec:appx-binding-intervention}.
%  $\Delta$target isolates the change in the final output logit assigned to the counterfactual target answer itself. (iv)

\begin{table}[!h]
\centering
\resizebox{\columnwidth}{!}{
\begin{tabular}{lrrr}
\toprule
Condition & 
$\Delta R$ & 
$\Delta$logit &
% $\Delta$target &
Switch\\
\midrule

Random
& -0.28
& 0.48
% & 0.38
& 0.06 \\

\hlcell{Q int. ($\uparrow$)}
& \hlcell{\textbf{2.66}} 
& \hlcell{\textbf{4.25}} 
% & \hlcell{\textbf{4.24}}
& \hlcell{\textbf{0.47}} \\

\hlcell{K int. (box+obj) ($\uparrow$)}
& \hlcell{\textbf{2.08}} 
& \hlcell{\textbf{3.51}} 
% & \hlcell{\textbf{3.28}}
& \hlcell{\textbf{0.31}} \\

K int. (box) ($\uparrow$)
& 1.92
& 0.04
% & 0.49
& 0.42 \\

K int. (obj) ($\uparrow$)
& 2.25
& 4.35
% & 3.98
& 0.35 \\

\hlcell{Q+K int. (box+obj) ($\downarrow$)}
& \hlcell{\textbf{-0.12}} 
& \hlcell{\textbf{1.57}}  
% & \hlcell{\textbf{1.37}}
& \hlcell{\textbf{0.12}} \\

\bottomrule
\end{tabular}
}
\caption{
Q/K binding-ID interventions on Group B in Gemma-9B. The random control uses a matched norm random direction. Gray cells mark the diagnostic interventions.
}
\label{tab:qk-binding-intervention}
\end{table}

\paragraph{Results.}The Q/K binding interventions indicate that Gemma-9B selects addresses via binding ID matching (Table~\ref{tab:qk-binding-intervention}). Query side intervention strongly redirects attention toward the counterfactual binding ID ($\Delta R=2.66$), with corresponding behavioral shifts ($\Delta\mathrm{logit}=4.25$, switch fraction $0.47$). Key side interventions also redirect attention, especially when both box and object keys are shifted ($\Delta R=2.08$), while object only key shifts produce the largest logit change ($\Delta\mathrm{logit}=4.35$). In contrast, box only key shifts affect attention but not logits, suggesting that box tokens support address selection whereas object tokens are more directly coupled to readout. Crucially, the combined Q+K intervention cancels the effect ($\Delta R=-0.12$, switch fraction $0.12$), close to the random direction control. Overall, Gemma-9B appears to implement address selection through Q/K binding ID matching, with object side key vectors most directly driving downstream answer readout.

% Overall, these results provide stronger evidence for a shared linear Q/K binding-ID mechanism in Gemma-9B than in Llama-8B. In Llama-8B, Group B is sensitive to key-side binding manipulations, but the weak query-side effect and lack of Q+K cancellation suggest either a more key dominated pointer or a query side binding representation that is not well captured by the mean difference direction used for the intervention. Gemma-12B and Llama-3B show a similar pattern (Appendix~\ref{sec:appx-binding-intervention}).

% \begin{figure}[!h]
%     \centering
%     \includegraphics[width=0.5\textwidth]{./figures/info_type_path.pdf}
%     \caption{Rebinding mechanism information flow: binding like information flows to }
%     \label{fig:info_type_path}
% \end{figure}

% \subsection{Model difference}

% gemma vs. llama

% Gemma models use Grouped-query attention + interleaved local/global sliding window attention (layer-wise locality constraints). -- so the information route can be more narrow

% Llama models use Grouped-query attention. -- so the information can be more distributed

% + Further analysis for box-only, object-only experiments.
\section{Generalization of local state rebinding across models}

We next examine whether retrieval conditioned rebinding reflects a general mechanism across Gemma-12B, Llama-3B, Llama-8B. We first use path patching to identify candidate rebinding circuit at the attention head level (Section~\ref{sec:path-patching}). Then, we localize the pointer like heads that route binding information (Section~\ref{sec:retrieval-routing-group}) and perform binding ID interventions on these head groups (Section~\ref{sec:bid-intervention}).

\paragraph{Rebinding circuit evaluation.}Table~\ref{tab:entity_tracking_acc} shows that the pruned rebinding circuits recover much of the full model behavior across models. The results with additional evaluation metrics are reported in Appendix Table~\ref{tab:circuit_eval_other_metrics}.

\begin{table}[!h]
\centering
\resizebox{1\columnwidth}{!}{%
\begin{tabular}{@{}l c c c c@{}}
\toprule
% & \multicolumn{3}{c}{Accuracy} \\
% \cmidrule(lr){2-4}
Model & $F(M)$ & $F(Cir)$ & $F(Random)$ & \# Component \\
\midrule
Gemma-9B & 1    & 0.89 & 0.34 & 38 / 672 \\
Gemma-12B & 1    & 0.91 & 0.38 & 83 / 768 \\
Llama-3B  & 0.71 & 0.71 & 0.34 & 42 / 672 \\
Llama-8B  & 0.99 & 0.84 & 0.38 & 40 / 1024 \\
\bottomrule
\end{tabular}%
}
\caption{Full model and rebinding circuit accuracy across Gemma-\{9B, 12B\}, Llama-\{3B, 8B\} with random baseline with number of attention heads retained. Additional experiments restricted to examples correctly answered by Llama-3B are presented in Appendix~\ref{sec:llama3b-correct-only}.}
\label{tab:entity_tracking_acc}
\end{table}

\begin{figure*}[!h]
    \centering
    \begin{subfigure}[t]{0.3\textwidth}
        \centering
        \includegraphics[width=\linewidth]{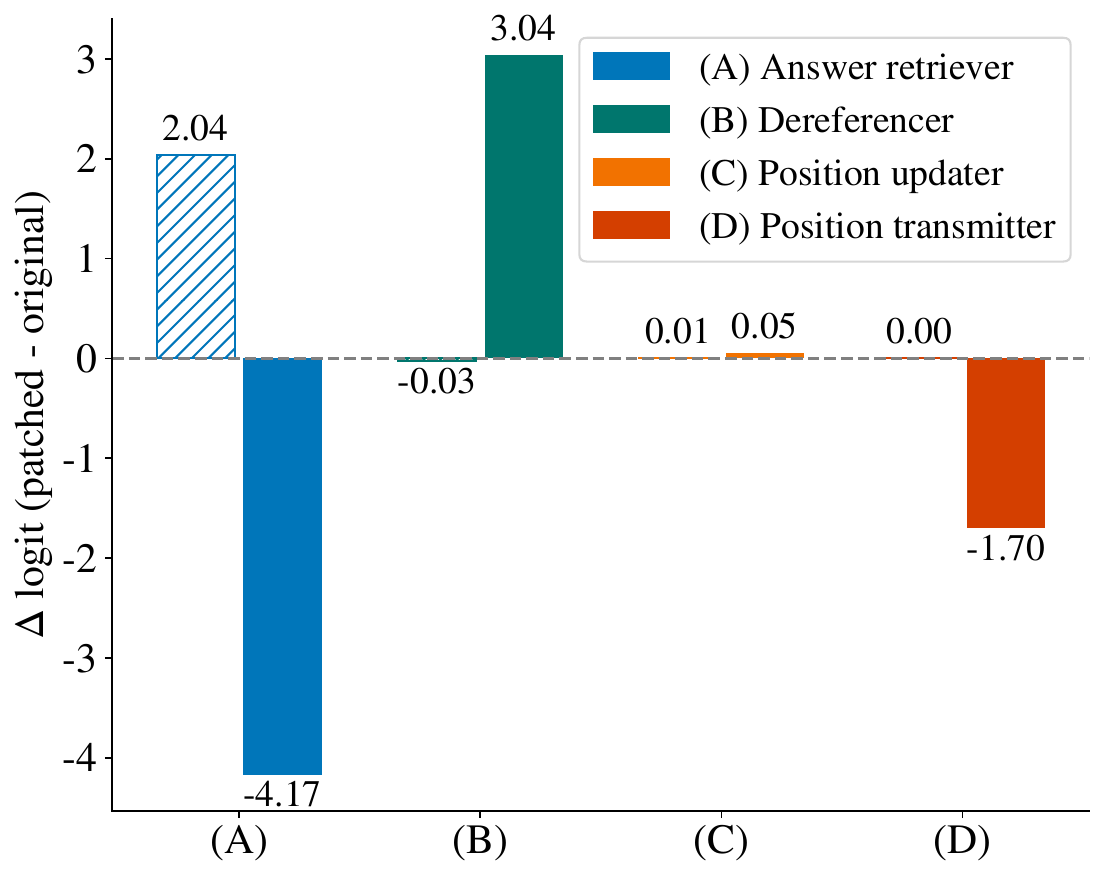}
        \caption{Gemma-12B}
        \label{fig:head_role_gemma12b}
    \end{subfigure}
      \begin{subfigure}[t]{0.3\textwidth}
        \centering
        \includegraphics[width=\linewidth]{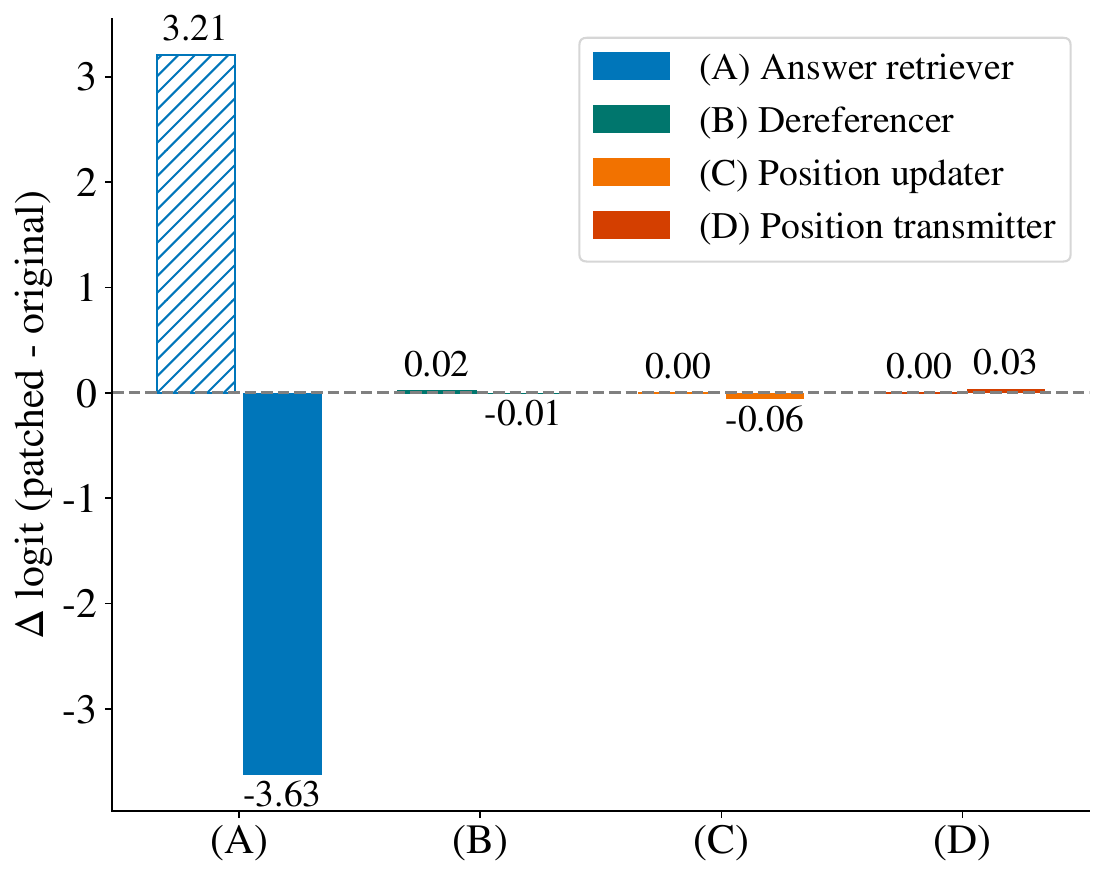}
        \caption{Llama-3B}
        \label{fig:head_role_llama3b}
    \end{subfigure}
    \begin{subfigure}[t]{0.3\textwidth}
        \centering
        \includegraphics[width=\linewidth]{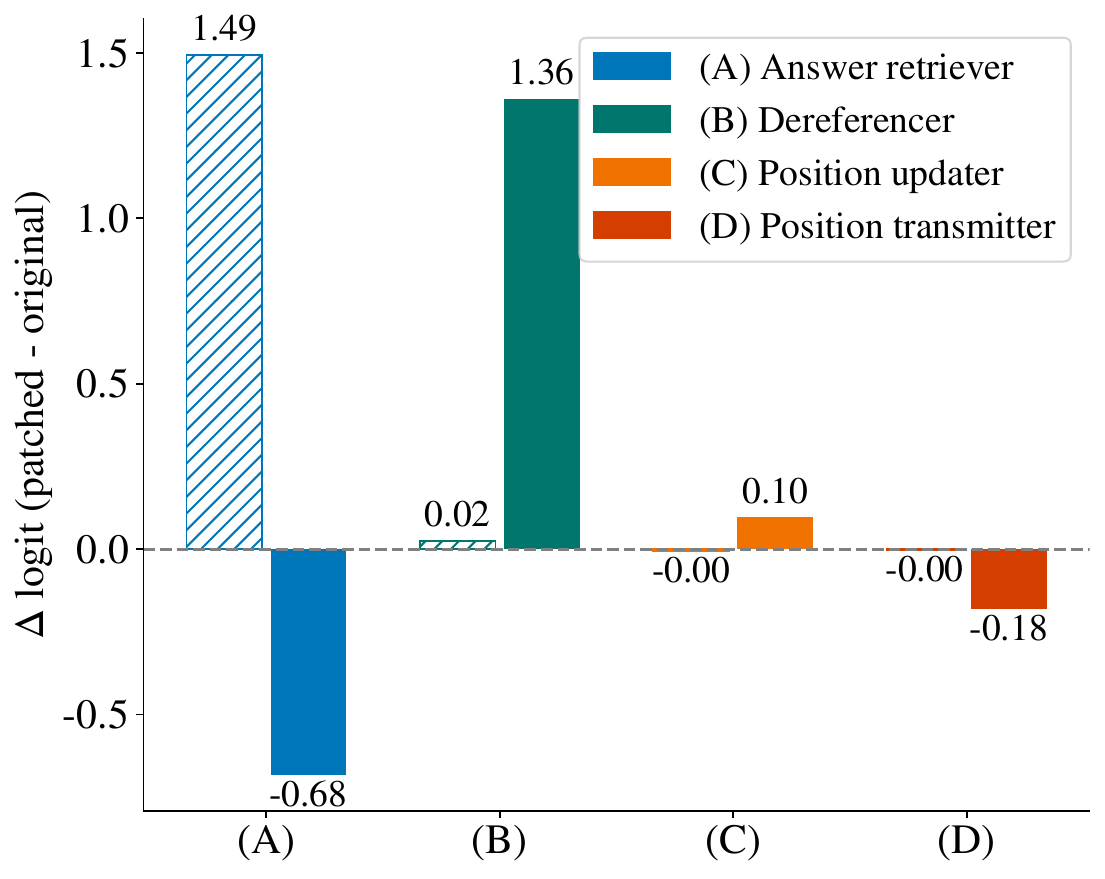}
        \caption{Llama-8B}
        \label{fig:head_role_llama8b}
    \end{subfigure}

    \caption{Role specific head patching effects across other models. 
\gemmabox~$\Delta \mathrm{logit}(\texttt{rabbit})$ is represented by fully colored bars, while \llamabox~$\Delta \mathrm{logit}(\texttt{egg})$ is represented by diagonally hatched bars.}
    \label{fig:head_role}
\end{figure*}

\paragraph{Functional roles of routing heads.}
\label{sec:function-head-other-models}
We next ask whether the head groups identified in the rebinding circuit play the same functional roles across models. As in Figure~\ref{fig:head_role}, across models (except Llama-3B), Group B remains important for address or pointer routing, where interchanging this group increases the $\Delta \mathrm{logit}$(\texttt{rabbit}), while leaving $\Delta \mathrm{logit}$(\texttt{egg}) comparatively unchanged. In contrast, Group A plays a larger role in answer object retrieval in the other models, where interchange on Group A increase the $\Delta \mathrm{logit}$(\texttt{egg}) while decreasing the $\Delta \mathrm{logit}$(\texttt{rabbit}). These results suggest a similar functional decomposition across models, with Group B primarily supporting address like routing and Group A contributing more directly to payload or answer retrieval.

\paragraph{Binding ID matching Q/K intervention.}The intervention results show that the binding ID mechanism generalizes across model families, but with different Q/K-level implementations. As shown in Table~\ref{tab:qk-binding-intervention-other-models}, Gemma-12B preserves the weak Q/K matching pattern observed in Gemma-9B (query side intervention produces the strongest attention and logit shifts ($\Delta R=0.26$, $\Delta_{\mathrm{logit}}=0.41$), while Q+K intervention cancels the effect ($\Delta R=-0.04$, $\Delta_{\mathrm{logit}}=-0.49$)). In contrast, Llama models are more key dominated. That is, key side interventions produce the largest attention shifts, query side effects are weak, and Q+K intervention remains positive ($\Delta R=1.69$ for Llama-3B; $\Delta R=0.98$ for Llama-8B). Thus, local state rebinding appears across architectures, but Gemma models exhibit clearer Q/K matching whereas Llama models rely more on key side binding information.

\begin{table}[!h]
\centering
\footnotesize
\setlength{\tabcolsep}{2.5pt}
\renewcommand{\arraystretch}{1.05}

\resizebox{\columnwidth}{!}{%
\begin{tabular}{@{}llrrr@{}}
\toprule
Model & Condition &
$\Delta R$ &
$\Delta$logit &
% $\Delta$target &
Switch\\
\midrule

\multirow{6}{*}{Gemma-12B}
& Random
& -0.12 & -0.69 
% & -0.27 
& 0.08 \\

& \hlcell{Q int. ($\uparrow$)}
& \hlcell{\textbf{0.26}}
& \hlcell{\textbf{0.41}}
% & \hlcell{\textbf{0.52}}
& \hlcell{\textbf{0.04}} \\

& \hlcell{K int. (box+obj) ($\uparrow$)}
& \hlcell{\textbf{0.29}}
& \hlcell{\textbf{0.10}}
% & \hlcell{\textbf{0.31}}
& \hlcell{\textbf{0.14}} \\

& K int. (box) ($\uparrow$)
& -0.33 & -0.05 
% & -0.03 
& 0.12 \\

& K int. (obj) ($\uparrow$)
& 0.87 & 0.23 
% & 0.67 
& 0.18 \\

& \hlcell{Q+K int. (box+obj) ($\downarrow$)}
& \hlcell{\textbf{-0.04}}
& \hlcell{\textbf{-0.49}}
% & \hlcell{\textbf{-0.40}}
& \hlcell{\textbf{0.08}} \\

\midrule

\multirow{6}{*}{Llama-3B}
& Random
& -0.10 & 0.21 
% & -0.02 
& 0.05 \\

& \hlcell{Q int. ($\uparrow$)}
& \hlcell{\textbf{-0.46}}
& \hlcell{\textbf{0.15}}
% & \hlcell{\textbf{0.06}}
& \hlcell{\textbf{0.01}} \\

& \hlcell{K int. (box+obj) ($\uparrow$)}
& \hlcell{\textbf{1.32}}
& \hlcell{\textbf{1.03}}
% & \hlcell{\textbf{0.42}}
& \hlcell{\textbf{0.30}} \\

& K int. (box) ($\uparrow$)
& 2.23 & 0.06
% & 0.04 
& 0.37 \\

& K int. (obj) ($\uparrow$)
& 0.29 & 0.69 
% & 0.34 
& 0.15 \\

& \hlcell{Q+K int. (box+obj) ($\downarrow$)}
& \hlcell{\textbf{1.69}}
& \hlcell{\textbf{0.95}}
% & \hlcell{\textbf{0.37}}
& \hlcell{\textbf{0.34}} \\
\midrule

\multirow{6}{*}{Llama-8B}
& Random
& 0.17 & 1.02 
% & 0.45 
& 0.12  
\\

& \hlcell{Q int. ($\uparrow$)}
& \hlcell{\textbf{0.18}} 
& \hlcell{\textbf{0.83}} 
% & \hlcell{\textbf{0.68}} 
& \hlcell{\textbf{0.06}}  
\\

& \hlcell{K int. (box+obj) ($\uparrow$)}
& \hlcell{\textbf{1.20}} 
& \hlcell{\textbf{2.92}} 
% & \hlcell{\textbf{2.15}} 
& \hlcell{\textbf{0.36}} 
\\

& K int. (box) ($\uparrow$)
& 1.59 & -0.02 
% & 0 
& 0.40  
\\

& K int. (obj) ($\uparrow$)
& 0.78
& 2.70 
% & 1.87 
& 0.26  
\\

& \hlcell{Q+K int. (box+obj) ($\downarrow$)}
& \hlcell{\textbf{0.98}} 
& \hlcell{\textbf{1.05}} 
% & \hlcell{\textbf{0.79}} 
& \hlcell{\textbf{0.27}} 
\\

\bottomrule
\end{tabular}%
}
\caption{Q/K binding ID interventions on Group B. The random control uses a matched norm random direction. \(\Delta R\): change in log attention ratio toward the counterfactual over the original target; $\Delta$logit: corresponding change in the counterfactual original object logit difference; Switch: switch rate of top attended context token from \(i\) to \(j\). Gray cells mark the diagnostic Q-side, K-side and Q+K cancellation interventions.}
\label{tab:qk-binding-intervention-other-models}
\end{table}

% The Q/K binding ID reveal model dependent patterns as in Table~\ref{tab:qk-binding-intervention-other-models}. Gemma-12B shows a weak Q/K matching signature observed in Gemma-9B. That is, the query side intervention produces the strongest attention and logit shifts ($\Delta R=0.26$, $\Delta_{\mathrm{logit}}=0.41$), while key side interventions have weaker or negative effects. Also, the combined Q+K intervention cancels the effect ($\Delta R=-0.04$, $\Delta_{\mathrm{logit}}=-0.49$). In contrast, the Llama models are more key side dominated, where key interventions produce the largest attention shifts, whereas query interventions are weak. Moreover, the combined Q+K intervention does not cancel the effect ($\Delta R=1.69$ for Llama-3B; $\Delta R=0.98$ for Llama-8B). Thus, while local state rebinding appears across models, its Q/K-level implementation differs. Gemma models show a clearer Q/K matching pattern, whereas Llama models rely more strongly on key side binding information.

\section{Conclusion}
We investigate how LLMs perform dynamic entity tracking in natural language contexts. Our findings support a mechanistic account in which LLMs do not simply reconstruct a complete updated world state after each state changing operation. Instead, they combine abstract binding representations at retrieval time and perform pointer updates. That is, the model preserves the original object representations and uses swap related information to redirect retrieval to the appropriate binding ID at retrieval time. This mechanism provides a concrete circuit level explanation for context dependent retrieval in transformers. 

% Our analyses show how role specific information is routed through token positions, how  a compact set of attention heads supports rebinding, and how modifying binding identifiers can redirect retrieval toward the corresponding object. These findings suggest that LLMs can solve dynamic tracking tasks through an efficient structured strategy by maintaining reusable binding information while applying local remapping only when the answer is retrieved. 

At the same time, the mechanism is not entirely identical across models. Although Gemma and Llama models show evidence for retrieval conditioned rebinding, they differ in functional roles of the attention head groups and Q/K signatures associated with the relevant heads. This suggests that the rebinding mechanism may be a shared computational strategy, but its circuit level implementation varies across model families. Understanding these cross-model differences is an important direction for future work, particularly for clarifying which aspects of rebinding reflect model architecture, scale, or properties of the pretraining distribution.

% We identify this rebinding mechanism in two steps. First, we locate the token positions that are causally important for the final prediction. Second, we localize an attention head level circuit over these positions using path patching. We show that a small subset of attention heads, comprising only 3--10\% of all heads, can recover performance comparable to that of the full model on the task. We further analyze the role of each circuit component and find that, in Gemma-9B, pointer like updating is primarily mediated by a dereferencer attention head group that uses binding IDs encoded in the Q/K subspace.

% In sum, our findings support a mechanistic account in which large language models handle dynamic state updates by composing stable binding representations with local pointer like updates. This offers a more precise explanation of how entity tracking can succeed without requiring full reconstruction of the latent world state, and it contributes to a mechanistic understanding of how transformer circuits resolve retrieval tasks.

\section*{Limitations}
While our task setting enables precise causal intervention, it is highly controlled and may not capture the full range of mechanisms used in natural discourse. Entities are represented as alphabetically named boxes, objects are single tokens, and state changes are limited to a single swap operation. Moreover, although the discovered circuits preserve much of the full model performance, they may omit redundant or backup components. Future work should test whether the same circuit structure generalizes to multiple swaps, larger entity sets, other state transitions such as put and remove, and more naturalistic narratives.

% Second, although the experiments cover several LLMs, the analysis remains limited. The results therefore do not establish that reference conditioned rebinding is universal across architectures. In particular, the Q/K binding ID intervention results vary across models: Gemma-9B shows a relatively clear Q/K binding matching signature, while Llama-8B, Llama-3B, and Gemma-12B show weaker or more key-side dominated effects. This suggests that the proposed mechanism may have unknown variants (such as architecture or training dependent).

Additionally, the binding ID directions used in the Q/K intervention experiments are estimated with mean difference probes under an approximate linear representation assumption. If binding information is represented nonlinearly, distributed across multiple subspaces, or dynamically transformed across layers, these interventions may underestimate the relevant mechanism.

\section*{Ethical Considerations}
All experiments were conducted on a single NVIDIA H100 PCIe GPU with 80 GB of memory with synthetic prompts involving boxes, objects, and swap operations. This work therefore poses minimal risks of data privacy and human subjects, although its GPU based analyses can incur the environmental costs. Specifically, the most computationally intensive experiment was path patching with Gemma-12B required $\sim$16 hours to complete. The intervention methods are also potentially dual-use, as they can be used both to diagnose and to manipulate model behavior.

% This work explores the internal mechanisms of dynamic entity tracking in large language models. The experiments are conducted on synthetic prompts involving boxes, objects, and swap operations. Therefore, the direct risks associated with data privacy and human subject participation are minimal.

\section*{Acknowledgements}
This project has received funding from the European Research Council (ERC) under the European Union's
Horizon 2020 research and innovation programme (ERC Starting Grant “Individualized Interaction in Discourse”,
grant agreement No. 948878). We gratefully acknowledge the stimulating research environment of the GRK 2853/1 ``Neuroexplicit Models of Language, Vision, and Action'', funded by the Deutsche Forschungsgemeinschaft (DFG, German Research Foundation) under project number 471607914.

\begin{figure}[h] 
\centering
\includegraphics[width=0.75\columnwidth]{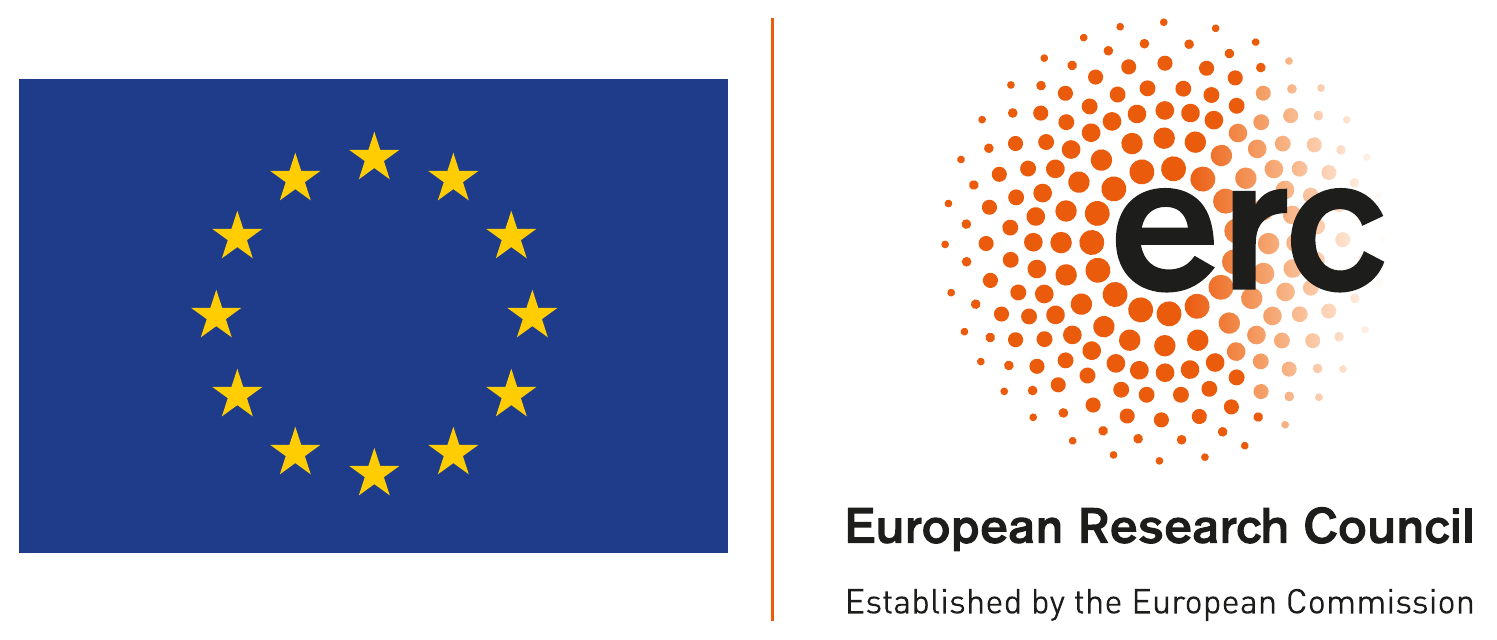}
\end{figure}

\bibliography{custom}

\appendix
\section{Causal mediation analysis for tracing information flow}
\label{sec:appx-causal-mediation}

\begin{figure}[!h]
    \centering
    \begin{subfigure}[t]{0.5\textwidth}
        \centering
        \includegraphics[width=\linewidth]{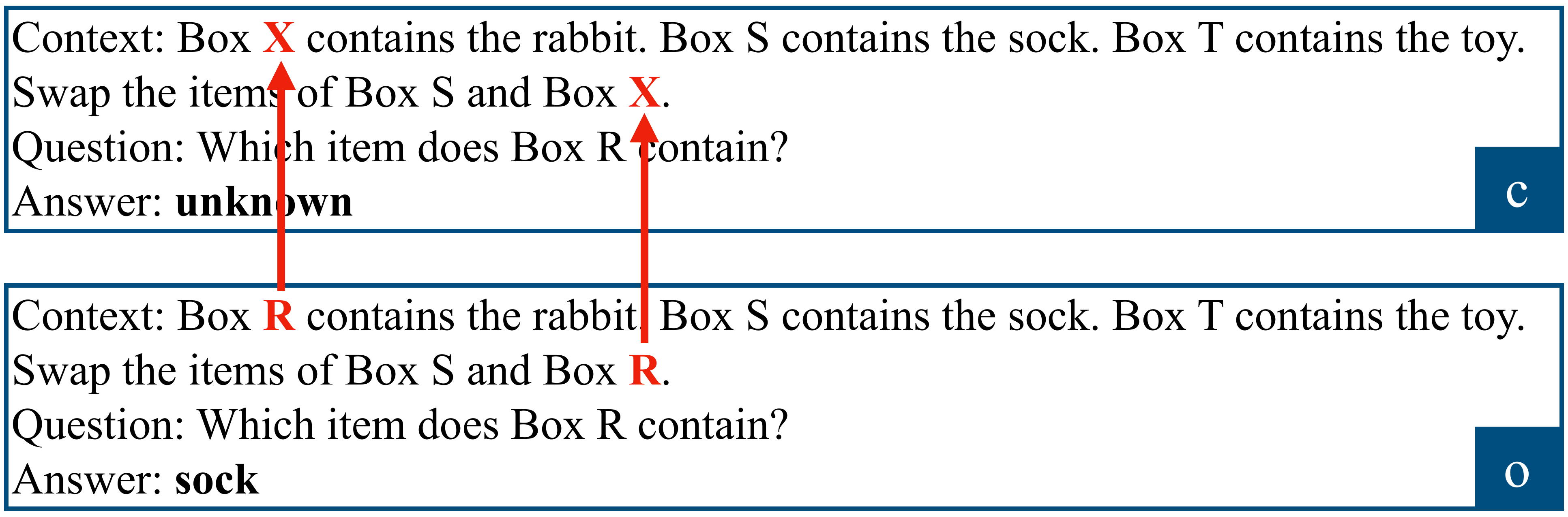}
        \includegraphics[width=\linewidth]{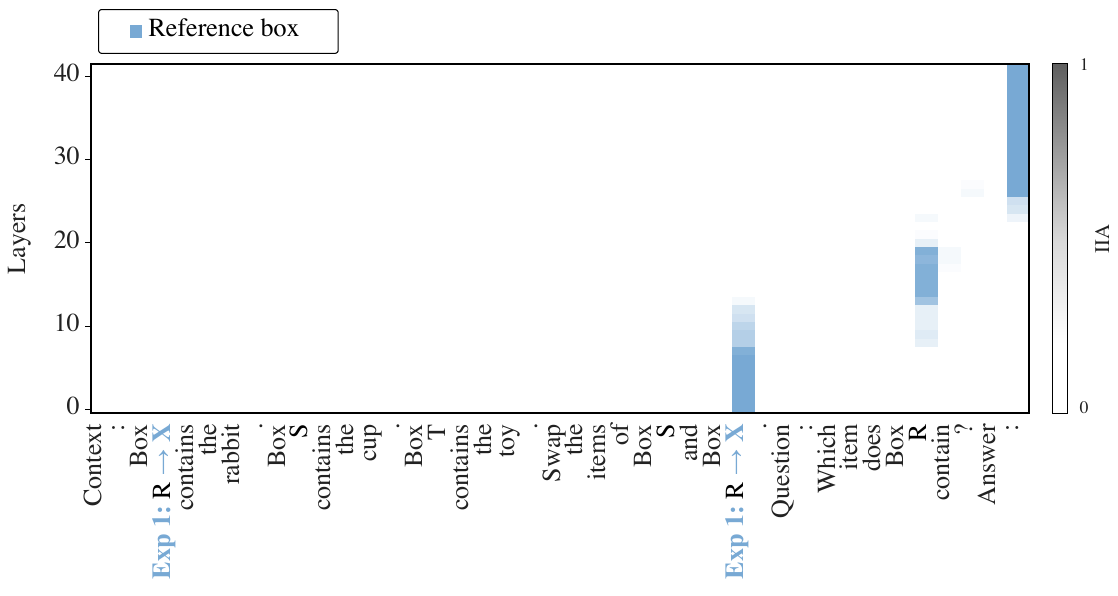}
        \caption{Exp1. Reference box tokens.}
        \label{fig:cm_qbox}
    \end{subfigure}
      \begin{subfigure}[t]{0.5\textwidth}
        \centering
        \includegraphics[width=\linewidth]{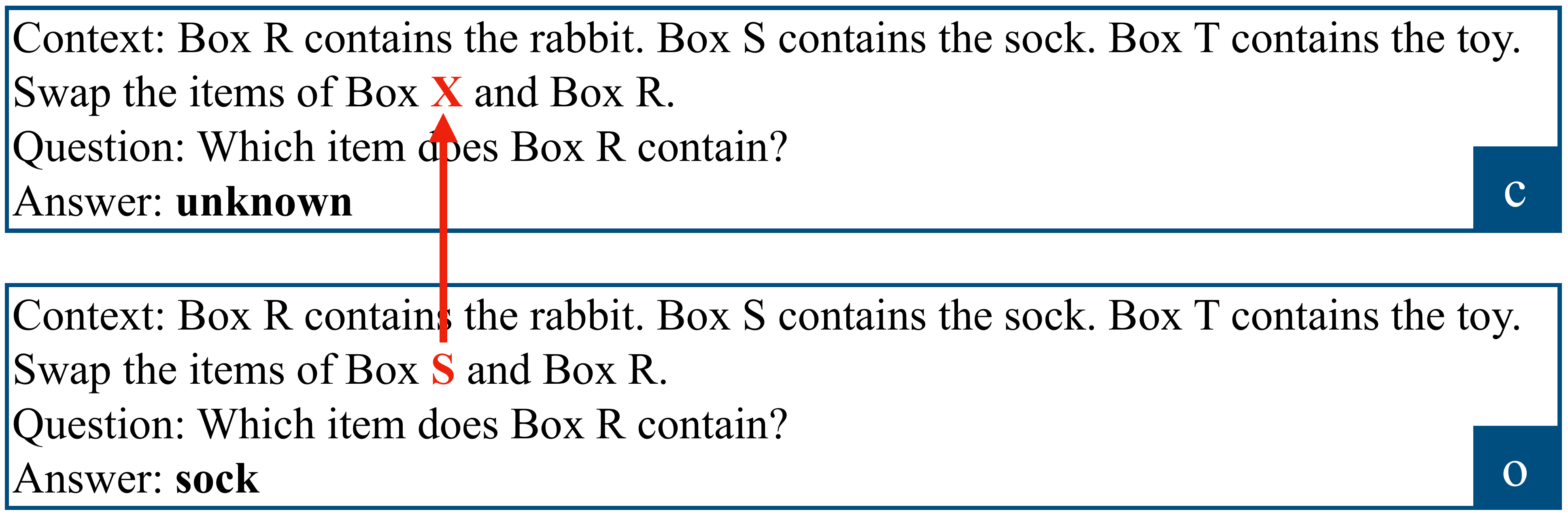}
        \includegraphics[width=\linewidth]
        {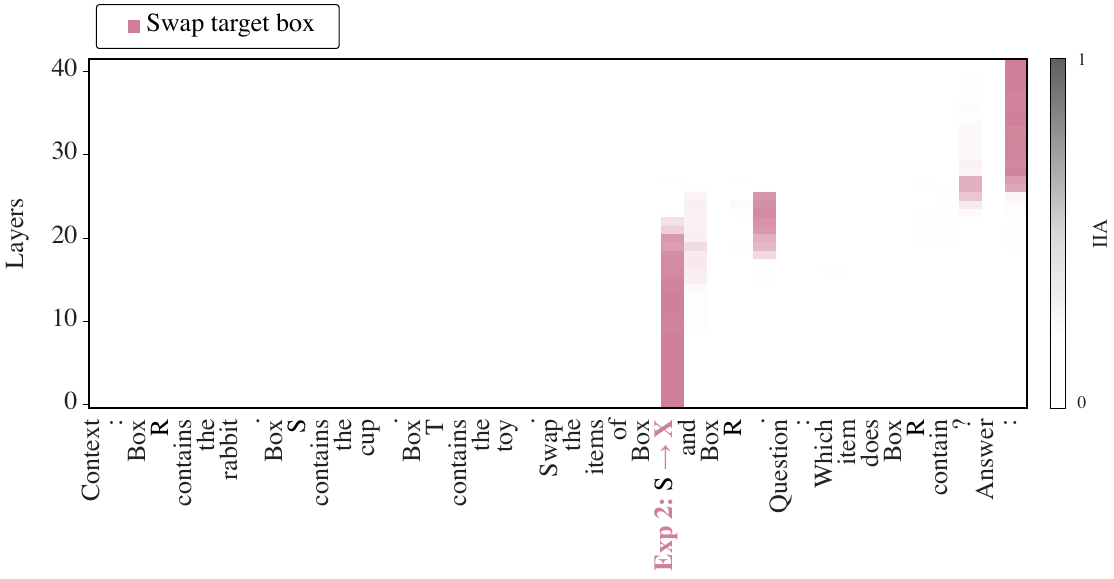}
        \caption{Exp2. Swap target box token.}
        \label{fig:cm_sbox}
    \end{subfigure}
    \begin{subfigure}[t]{0.5\textwidth}
        \centering
        \includegraphics[width=\linewidth]{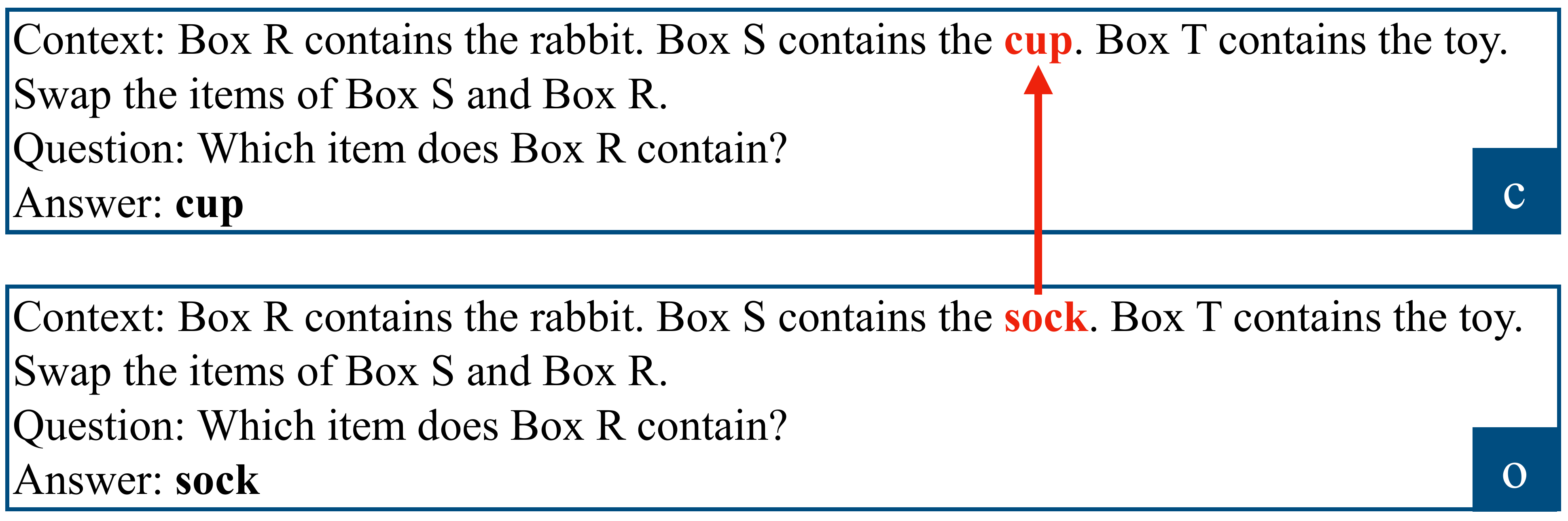}
        \includegraphics[width=\linewidth]{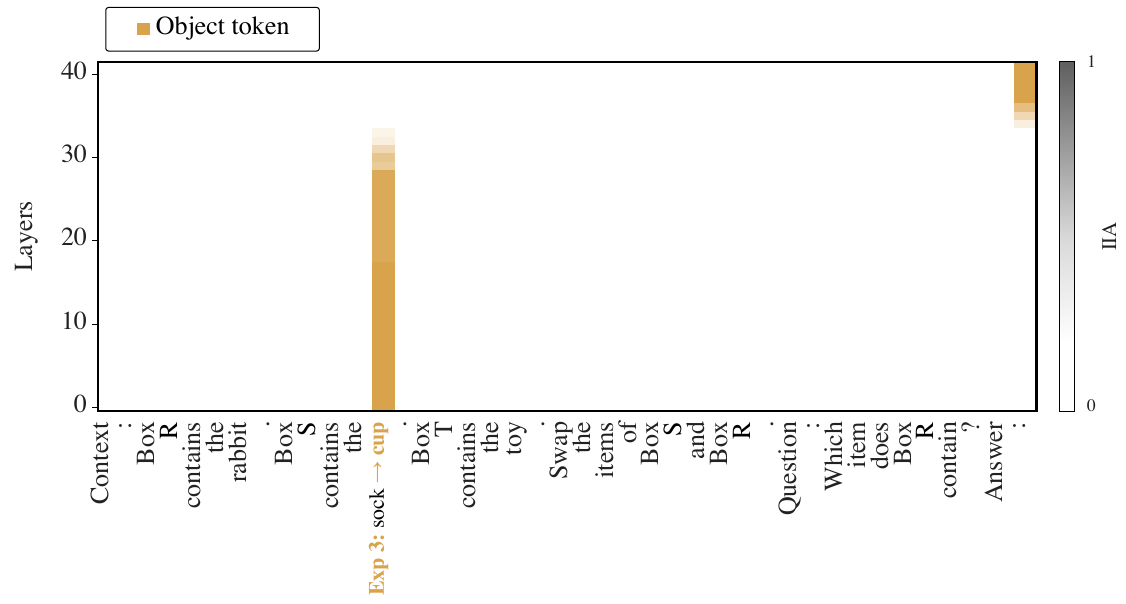}
        \caption{Exp3. Object token.}
        \label{fig:cm_o}
    \end{subfigure}
    \caption{Information flow via causal mediation analysis per-experiment setup and result.}
    \label{fig:cm}
\end{figure}

Figure~\ref{fig:cm} shows the causal mediation analysis setup and per-experiment results. 

\begin{figure*}[!h]
    \centering
    \includegraphics[width=0.8\textwidth]
    {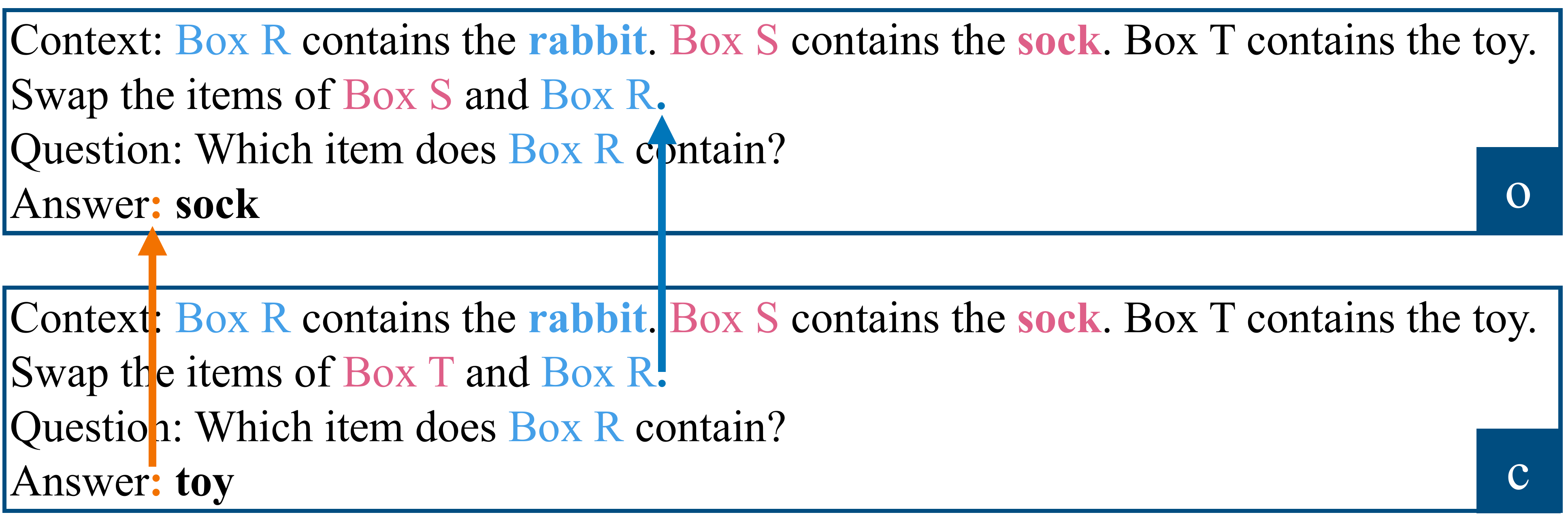}
    \includegraphics[width=1\textwidth]
    {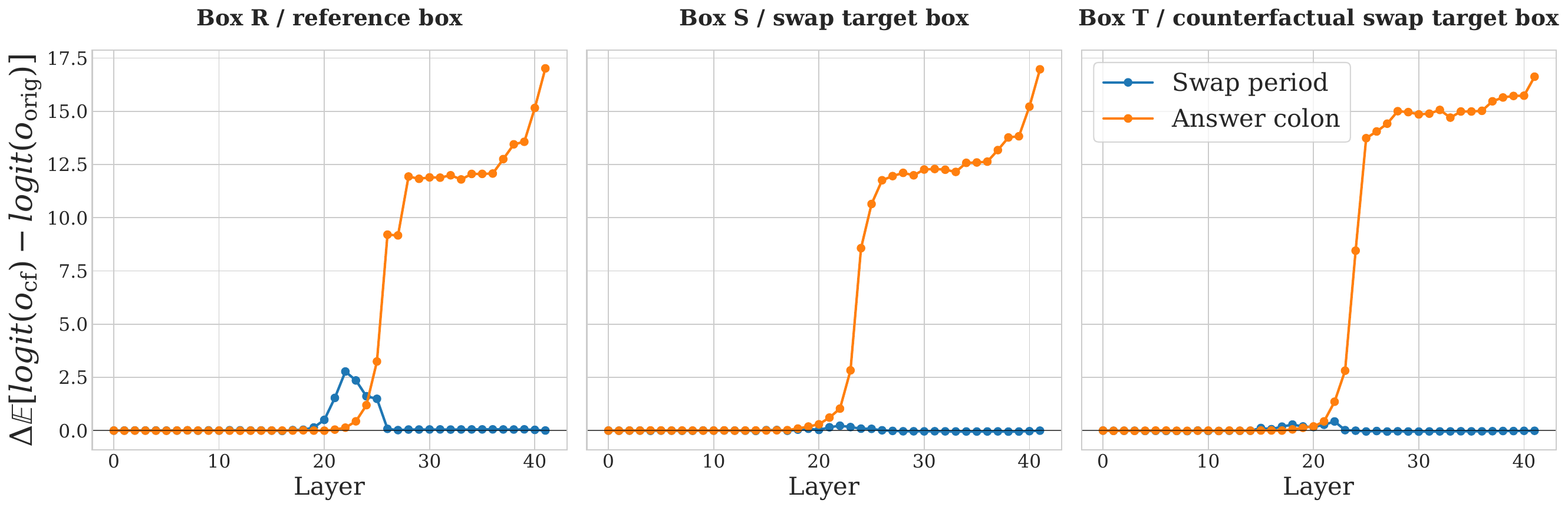}
 \caption{Additional causal mediation diagnostic comparing post swap prefix patching with readout position patching. \textbf{Top:} We patch activations from \(c\) into \(o\) either at the period following the swap instruction (\textcolor{mpblue}{post swap prefix patching; blue arrow}) or at the answer colon position (\textcolor{mporange}{readout patching; orange arrow}). \textbf{Bottom:} Change in mean counterfactual minus original logit margin relative to the clean baseline. Positive values indicate that the patch shifts the model toward the counterfactual answer.}
    \label{fig:cm_additional}
\end{figure*}

We additionally conduct experiment for patching the `.' position after the swap operation (post swap prefix) and compare it with when patching `:' position at the retrieval. If the post swap prefix already encodes the updated global state, then patching activations at post swap prefix positions should have an effect comparable to patching the activation at the final answer position. If rebinding is computed during retrieval, answer position patches should produce much stronger counterfactual transfer than prefix patches. We further repeat the same patch but questions across all boxes (i.e., Which item does Box \{b\} contain?, $b \in \{R,S,T\}$). If the same prefix patch transfers the whole counterfactual mapping for all three boxes (i.e., increase the logit for counterfactual mapping), then it indicates for global state update right after the swap operation. If it only helps a specific referenced box, then it is probably encoding the swap operation rather than a complete updated state. As in Figure~\ref{fig:cm_additional}, patching the post swap prefix produces only a localized counterfactual effect on the referenced box, whereas patching the answer colon position yields a stronger and more consistent effect across all the boxes. This pattern suggests that swap related information is present before the question, but the relevant binding is resolved mainly during retrieval rather than stored as a complete post swap state. 

\subsection{Generalization beyond the three-box setting}
\label{sec:generalization-threebox}

\begin{figure*}[!h]
    \centering
    \includegraphics[width=1\textwidth]{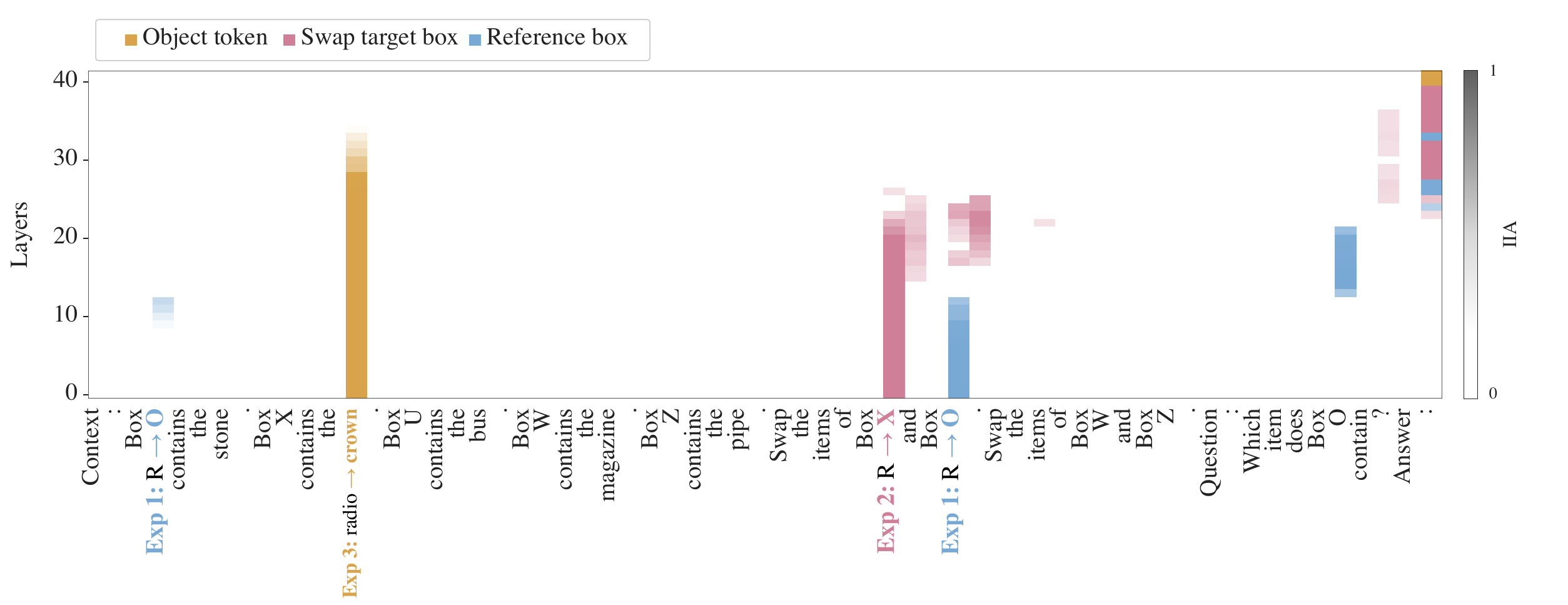}
 \caption{Information flow of crucial tokens using interchange interventions in Gemma-9B with larger number of boxes (N=5) and an additional swap distractor. \textbf{y-axis:} model layers; \textbf{x-axis:} shared original prompt template. Each colored token is varied in a separate experiment: \textcolor{qbox}{Exp1. Reference box}, \textcolor{pbox}{Exp2. Swap target box}, \textcolor{object}{Exp 3. Object}, all other black tokens are unchanged; \textbf{cell:} reports normalized IIA, darker cells indicating stronger causal mediation.}
    \label{fig:iia_5boxes}
\end{figure*}

\begin{figure*}[!h]
    \centering
    \includegraphics[width=1\textwidth]
    {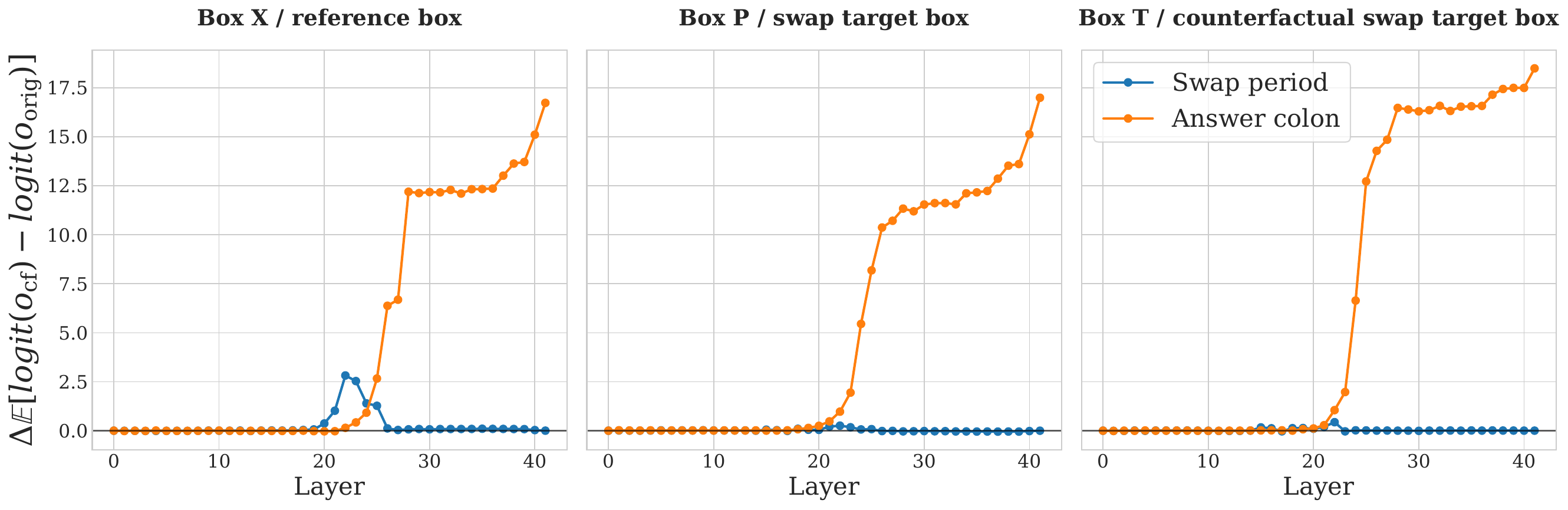}
 \caption{Additional causal mediation diagnostic comparing post swap prefix patching with readout position patching with 5 boxes and a distractor swap operation. Change in mean counterfactual minus original logit margin relative to the clean baseline. Positive values indicate that the patch shifts the model toward the counterfactual answer.}
    \label{fig:cm_additional_5boxes}
\end{figure*}

We extended these analyses to a setting with more boxes and an additional distractor swap, as shown in Figure~\ref{fig:iia_5boxes} and Figure~\ref{fig:cm_additional_5boxes}.

% \begin{figure*}[!h]
%     \centering
%     \includegraphics[width=1\textwidth]{./figures/appx/combined_overlay_iia_5boxes.pdf}
%  \caption{Information flow of crucial tokens using interchange interventions in Gemma-9B with larger number of boxes (N=5) and an additional swap distractor. \textbf{y-axis:} model layers; \textbf{x-axis:} shared original prompt template. Each colored token is varied in a separate experiment: \textcolor{qbox}{Exp1. Reference box}, \textcolor{pbox}{Exp2. Swap target box}, \textcolor{object}{Exp 3. Object}, all other black tokens are unchanged; \textbf{cell:} reports normalized IIA, darker cells indicating stronger causal mediation.}
%     \label{fig:iia_5boxes}
% \end{figure*}

% \begin{figure*}[!h]
%     \centering
%     \includegraphics[width=1\textwidth]
%     {./figures/appx/side_by_side_per_question_comparison_with_question_5boxes.pdf}
%  \caption{Additional causal mediation diagnostic comparing post swap prefix patching with readout position patching with 5 boxes and a distractor swap operation. Change in mean counterfactual minus original logit margin relative to the clean baseline. Positive values indicate that the patch shifts the model toward the counterfactual answer.}
%     \label{fig:cm_additional_5boxes}
% \end{figure*}

\section{Selected attention heads in rebinding circuit}
\label{sec:circuit-heads}

Table~\ref{tab:circuit_eval_other_metrics} shows the mean candidate margin and mean label logit of the full model, pruned circuit, and random baseline. Table~\ref{tab:attn_heads_four_models} reports the attention heads selected after pruning for each group in the rebinding circuit. 

% After pruning, some models retain no heads in particular groups, suggesting that the functional decomposition may be implemented redundantly or through model specific variants.

\begin{table}[!h]
\centering
\setlength{\tabcolsep}{3.5pt}
\resizebox{1\columnwidth}{!}{%
\begin{tabular}{@{}l ccc ccc@{}}
\toprule
& \multicolumn{3}{c}{Mean candidate margin}
& \multicolumn{3}{c}{Mean label logit} \\
\cmidrule(lr){2-4}
\cmidrule(lr){5-7}
Model
& M & Cir & Rand.
& M & Cir & Rand. \\
\midrule
Gemma-9B
& 8.87 & 2.76 & -0.57
& 20.33 & 12.35 & 7.67 \\

Gemma-12B
& 17.83 & 14.10 & -0.78
& 31.73 & 21.69 & 6.22 \\

Llama-3B
& 0.84 & 2.32 & -0.80
& 20.24 & 16.36 & 11.49 \\

Llama-8B
& 4.72 & 3.70 & -0.52
& 20.77 & 14.72 & 9.59 \\
\bottomrule
\end{tabular}%
}
\caption{
Additional circuit evaluation metrics across models. We report candidate margin and mean label logit for the full model (M), the pruned rebinding circuit (Cir), and random head subsets of matched size (Rand.) on held out examples ($N=300$).
}
\label{tab:circuit_eval_other_metrics}
\end{table}

\section{Functional roles of attention heads groups}
\label{sec:appx-attention-head-ii}

Appendix Figure~\ref{fig:head_role} provides the full per-model results for the head role interventions as described in Section~\ref{sec:function-head-other-models}. 

% \textcolor{red}{We additionally conduct experiment on attention head group (C) Position updater. We assumed that the attention group even though lexically it's aligned with the different box, but it will have higher attention score on the box that it should attend to have the updated binding ID from. Selected-head attention aggregate
% mean(Box R question -> context Box R): 0.003173
% mean(Box R question -> context Box S): 0.008113
% mean difference R-S: -0.004941}

% \begin{figure*}[!h]
%     \centering
%     \begin{subfigure}[t]{0.3\textwidth}
%         \centering
%         \includegraphics[width=\linewidth]{figures/appx/other_models/delta_head_role_gemma12b.pdf}
%         \caption{Gemma-12B}
%         \label{fig:head_role_gemma12b}
%     \end{subfigure}
%       \begin{subfigure}[t]{0.3\textwidth}
%         \centering
%         \includegraphics[width=\linewidth]{figures/appx/other_models/delta_head_role_llama3b.pdf}
%         \caption{Llama-3B}
%         \label{fig:head_role_llama3b}
%     \end{subfigure}
%     \begin{subfigure}[t]{0.3\textwidth}
%         \centering
%         \includegraphics[width=\linewidth]{figures/appx/other_models/delta_head_role_llama8b.pdf}
%         \caption{Llama-8B}
%         \label{fig:head_role_llama8b}
%     \end{subfigure}

%     \caption{Role specific head patching effects across other models. 
% \gemmabox~$\Delta \mathrm{logit}(\texttt{rabbit})$ is represented by fully colored bars, while \llamabox~$\Delta \mathrm{logit}(\texttt{egg})$ is represented by diagonally hatched bars.}
%     \label{fig:head_role}
% \end{figure*}

\section{Binding ID intervention metrics}
\label{sec:appx-binding-intervention}

% \subsection{Residual-stream binding ID intervention}
% \textcolor{red}{To locate where the model maintains or updates the binding ID, we further conduct an intervention experiment on the residual stream at token positions associated with the swap operation's boxes and the reference box at the question~\cite{feng2024language}. In this experiment, we extract binding vectors from the context box positions and then add or subtract these vectors at selected positions.} 

% \textcolor{red}{Our results suggest that the binding ID has been updated at the question box token position. The intervention at the swap operation positions lead to a substantial performance degradation, with accuracy to 0.48. This pattern is consistent with our other findings that during the swap operation, the model must preserve the relevant original binding ID for subsequent selection. Therefore, interchanging the binding ID at these positions disrupts the model's ability to maintain the correct binding. In contrast, intervening at the question box token position has a relatively small effect where the performance remains 0.83, compared with the no patching baseline of 1. This indicates that the question position either requires a binding ID update or already contains the updated binding ID therefore making the additional binding vector less disruptive.}

% \subsection{Metric definitions}
For each example, let $i$ denote the original binding ID and $j$ denote the counterfactual binding ID induced by the intervention. 
Let $S_k$ be the set of context token positions associated with binding ID $k$, and let $A_{\ell,h}(t_{\mathrm{:}},s)$ denote the attention weight from the relevant query position $t_{\mathrm{:}}$ to source position $s$ for head $(\ell,h)$. 
We define the attention mass assigned to binding ID $k$ as
\[
M_k = \sum_{s \in S_k} A_{\ell,h}(t_{\mathrm{:}},s).
\]

The change in log attention ratio is
\[
\begin{aligned}
\Delta R =
\left[
\log(M_j+\epsilon)-\log(M_i+\epsilon)
\right]_{\mathrm{patched}} \\
-
\left[
\log(M_j+\epsilon)-\log(M_i+\epsilon)
\right]_{\mathrm{original}},
\end{aligned}
\]
where $\epsilon$ is a small constant for numerical stability. 
A positive $\Delta R$ indicates that the intervention shifts attention from the original binding ID $i$ toward the counterfactual binding ID $j$.

Let $z(O)$ denote the final output logit assigned to object token $O$, and let $O_i$ and $O_j$ be the objects associated with the original and counterfactual binding IDs, respectively. 
We define the change in final logit margin as
\[
\begin{aligned}
\Delta{\mathrm{logit}}
=
\left[
z(O_j)-z(O_i)
\right]_{\mathrm{patched}} \\
-
\left[
z(O_j)-z(O_i)
\right]_{\mathrm{original}}.
\end{aligned}
\]

% We also report the change in the counterfactual target logit alone:
% \[
% \Delta{\mathrm{target}}
% =
% z_{\mathrm{patched}}(O_j)
% -
% z_{\mathrm{original}}(O_j).
% \]

Finally, the switch fraction measures the fraction of examples for which the
dominant attended binding ID changes from \(i\) in the original run to \(j\)
in the patched run. Let $o_n = \operatorname*{arg\,max}_{k} M^{(n)}_{k,\mathrm{original}},
p_n = \operatorname*{arg\,max}_{k} M^{(n)}_{k,\mathrm{patched}}$, then
\[
\mathrm{Switch}
=
\frac{1}{N}
\sum_{n=1}^{N}
\mathbf{1}\{o_n = i,\; p_n = j\}.
\]

\section{Binding ID intervention on Llama-3B only on correct answered examples}\label{sec:llama3b-correct-only}

Llama-3B results should be interpreted cautiously because it might be already weak on the task. However, we would not describe the found circuit accuracy as floor effect, where the circuit accuracy remains above the random head baseline (0.71; random: 0.34). We evaluated on the held-out dataset than the dataset that we found the circuit from to make sure that the found circuit is not just a random artifact. Moreover, we argue that the positive Q+K intervention for the Llama-3B model shows that Gemma model's binding Q/K mechanism does not hold for Llama model. We additionally conduct further experiment to address this point more directly.

We conduct additional head specific binding ID intervention only on the examples that the model answer correctly to get rid of the effect of the confounding error. As in Table~\ref{tab:correct_only_llama3b_int}, the model performs the similar as the original performance for the binding intervention where it uses all the examples. Therefore, the positive Q+K effect is probably not due to low task accuracy or failures. It does reflect the key-dominated mechanism.

\begin{table}[!h]
\centering
\resizebox{\columnwidth}{!}{%
\begin{tabular}{lrrrrrr}
\toprule
& \multicolumn{2}{c}{Q int.}
& \multicolumn{2}{c}{K int.}
& \multicolumn{2}{c}{Q+K int.} \\
\cmidrule(lr){2-3}
\cmidrule(lr){4-5}
\cmidrule(lr){6-7}
& $\Delta R$ & $\Delta\mathrm{logit}$
& $\Delta R$ & $\Delta\mathrm{logit}$
& $\Delta R$ & $\Delta\mathrm{logit}$ \\
\midrule
All
& $-0.46$ & $0.15$
& $1.32$  & $1.03$
& $1.69$  & $0.95$ \\
Only correct
& $-0.48$ & $0.11$
& $1.36$  & $1.05$
& $1.73$  & $1.05$ \\
\bottomrule
\end{tabular}%
}
\caption{Llama-3B intervention results for all vs. only corrected examples.}
\label{tab:correct_only_llama3b_int}
\end{table}

\begin{table*}[!h]
\centering
\small
\setlength{\tabcolsep}{3pt}
\renewcommand{\arraystretch}{1.08}
\begin{tabularx}{\textwidth}{@{}
>{\raggedright\arraybackslash}p{0.12\textwidth}
>{\raggedright\arraybackslash}p{0.18\textwidth}
>{\raggedright\arraybackslash}X
@{}}
\toprule
\textbf{Model} & \textbf{Category} & \textbf{Attention heads} $(h,\ell)$ \\
\midrule

\multirow{5}{*}{Gemma-9B}
& (A) Answer retriever &
(15, 27), (7, 25), (0, 26), (1, 25), (8, 39), (1, 39), (11, 40), (7, 21), (13, 27), (12, 38), (10, 25) \\

& (B) Dereferencer &
(12, 28), (4, 28), (1, 30), (5, 22), (14, 24), (3, 21), (8, 28), (10, 28), (6, 29), (12, 21), (12, 26), (7, 32), (15, 30), (4, 26), (4, 22), (15, 11), (2, 28) \\

& (C) Position updater &
(3, 18), (7, 21), (3, 19), (14, 19), (10, 16), (1, 18), (2, 0), (11, 22), (13, 8) \\

& (D) Swap position transmitter &
(0, 17) \\

& (E) Binding anchor &
- \\

\midrule

\multirow{5}{*}{Gemma-12B}
& (A) Answer retriever &
(2, 47), (1, 46), (14, 47), (4, 47), (10, 32), (7, 47), (4, 31), (13, 32), (2, 30), (0, 15), (15, 30), (4, 10), (8, 16), (12, 29), (5, 29), (11, 14), (3, 37), (1, 4), (13, 43) \\

& (B) Dereferencer &
(6, 28), (0, 46), (11, 32), (2, 44), (11, 41), (14, 1), (13, 6), (3, 18), (14, 30), (8, 46), (10, 6), (4, 26), (1, 9), (4, 39), (12, 26), (10, 15), (9, 6), (3, 0), (7, 45), (9, 46), (7, 26), (3, 41), (12, 44), (2, 46), (1, 35), (10, 28), (3, 30), (10, 3), (11, 46) \\

& (C) Position updater &
(6, 5), (8, 40), (2, 30), (0, 5), (14, 24), (4, 18), (6, 41), (9, 33), (2, 27), (15, 32), (14, 9), (14, 15), (13, 34), (8, 36), (9, 11), (10, 6), (14, 12), (11, 27) \\

& (D) Swap position transmitter &
(14, 7), (3, 10), (7, 19), (9, 9), (2, 35), (1, 19), (6, 13), (10, 5), (6, 28) \\

& (E) Binding anchor &
(8, 20), (13, 9), (7, 13), (10, 10), (11, 19), (6, 22), (6, 25), (4, 23) \\

\midrule

\multirow{5}{*}{Llama-3B}
& (A) Answer retriever &
(5, 24), (9, 18), (0, 27), (2, 15), (17, 21), (6, 16), (22, 15), (1, 27), (20, 21), (0, 17), (17, 16), (9, 23), (11, 22) \\

& (B) Dereferencer &
(15, 13), (2, 12), (1, 12), (5, 11), (23, 13), (17, 13), (18, 15), (3, 12), (11, 12), (16, 14), (14, 12), (19, 14), (10, 18), (6, 9), (3, 7), (7, 19), (16, 10), (7, 7), (2, 14), (1, 7) \\

& (C) Position updater &
(11, 10), (9, 11), (23, 9), (1, 11), (20, 9) \\

& (D) Swap position transmitter &
(22, 8), (19, 2), (7, 10), (2, 7) \\

& (E) Binding anchor &
- \\

\midrule

\multirow{5}{*}{Llama-8B}
& (A) Answer retriever &
(7, 27), (29, 22), (28, 16), (20, 18), (26, 17), (26, 16), (4, 17), (13, 24) \\

& (B) Dereferencer &
(6, 14), (13, 13), (10, 21), (11, 15), (1, 16), (13, 20), (8, 15), (5, 14), (24, 17), (15, 13), (21, 13), (28, 18), (3, 26), (16, 16), (5, 17), (3, 16), (19, 22), (25, 12), (9, 15), (11, 22), (25, 16), (26, 11), (13, 26), (2, 13), (26, 24) \\

& (C) Position updater &
(2, 10), (25, 12), (3, 10), (21, 13) \\

& (D) Swap position transmitter &
(13, 8), (0, 10) \\

& (E) Binding anchor &
(21, 5) \\

\bottomrule
\end{tabularx}
\caption{Attention heads remaining after pruning across four models. Heads are reported as $(h,\ell)$, where $h$ denotes the head index and $\ell$ denotes the layer index.}
\label{tab:attn_heads_four_models}
\end{table*}

\end{document}